\documentclass[10pt,twocolumn,letterpaper]{article}

\usepackage[pagenumbers]{cvpr} % To force page numbers, e.g. for an arXiv version

\usepackage{algorithm}
\usepackage{algpseudocode} 
\usepackage[table]{xcolor}

\definecolor{cvprblue}{rgb}{0.21,0.49,0.74}
\usepackage[breaklinks,colorlinks,citecolor=cvprblue]{hyperref}

\def\paperID{10474} % *** Enter the Paper ID here
\def\confName{CVPR}
\def\confYear{2024}

\title{Attention-Driven Training-Free Efficiency Enhancement of Diffusion Models}

\author{Hongjie Wang$^1$\thanks{Work was partly done during an internship at Adobe.}, Difan Liu$^2$, Yan Kang$^2$, Yijun Li$^2$, Zhe Lin$^2$, Niraj K. Jha$^1$, Yuchen Liu$^2$\thanks{Corresponding Author.}\\
$^1$Princeton University, $^2$Adobe Research
}

\begin{document}
\maketitle
\begin{abstract}
Diffusion Models (DMs) have exhibited superior performance in generating high-quality and diverse images. % with high quality and unparalleled diversity. 
However, this exceptional performance comes at the cost of expensive architectural design,
particularly due to the attention module heavily used in leading models.
Existing works mainly adopt a retraining process 
% under one specific compression ratio 
to enhance DM efficiency. This is computationally expensive and not very scalable.
%While there exist some previous works on enhancing the efficiency of DMs,
%most of them require a retraining process, which requires an enormous amount of computation, and very few of them look at the attention blocks, which have become the de facto inference bottleneck in the model.
% Most previous works that enhance the efficiency of diffusion models require an expensive retraining of the diffusion backbones. 
% Furthermore, if the targeted compression ratio is changed due to deployment on a different device, the retraining process needs to be repeated. 
% This strongly limits the application of these efficiency enhancement methods. 
To this end, we introduce the \textbf{A}ttention-driven \textbf{T}raining-free \textbf{E}fficient \textbf{D}iffusion \textbf{M}odel (\textbf{AT-EDM}) framework that leverages attention maps to perform run-time pruning of redundant tokens, without the need for any retraining. 
% We achieve this goal by reformulating the Weighted Page Rank (WPR) algorithm and converting attention maps to directed graphs to generate an importance score distribution for tokens. 
%Leveraging the rich information available from attention maps, 
Specifically, for single-denoising-step pruning, we develop a novel ranking algorithm, Generalized Weighted Page Rank (G-WPR), to identify redundant tokens, %where a graph algorithm is run to identify redundant tokens to accelerate attention computation 
and a similarity-based recovery method to restore tokens for the convolution operation.
In addition, we propose a Denoising-Steps-Aware Pruning (DSAP) approach to adjust the pruning budget across different denoising timesteps for better generation quality.
% We evaluate our method and baselines on the most powerful open-source text-to-image diffusion model, Stable Diffusion XL. 
Extensive evaluations show that AT-EDM performs favorably against prior art in terms of efficiency (e.g., 38.8\% FLOPs saving and up to 1.53$\times$ speed-up over Stable Diffusion XL) while maintaining nearly the same FID and CLIP scores as the full model. Project webpage: \href{https://atedm.github.io/}{https://atedm.github.io}.
%Compared with the state-of-the-art pruning method, our method achieves much better object reservation with sharper details and higher text-image alignment under the same FLOPs budget. 
%Notably, AT-EDM provides a 38.8\% off-the-shelf FLOPs saving on Stable Diffusion XL while maintaining nearly the same FID and CLIP scores. 
%as the full model.
%, outperforming prior arts.
%With fine-tuning, Edmea is able to reduce the FLOPs overhead by more than xx\% while keeping the quality of images.
\end{abstract}    
% \vspace{-2em}
\section{Introduction}
\label{sec:intro}

\begin{figure*}[t]
  \centering
  % \vspace{-2em}
   \includegraphics[width=\linewidth]{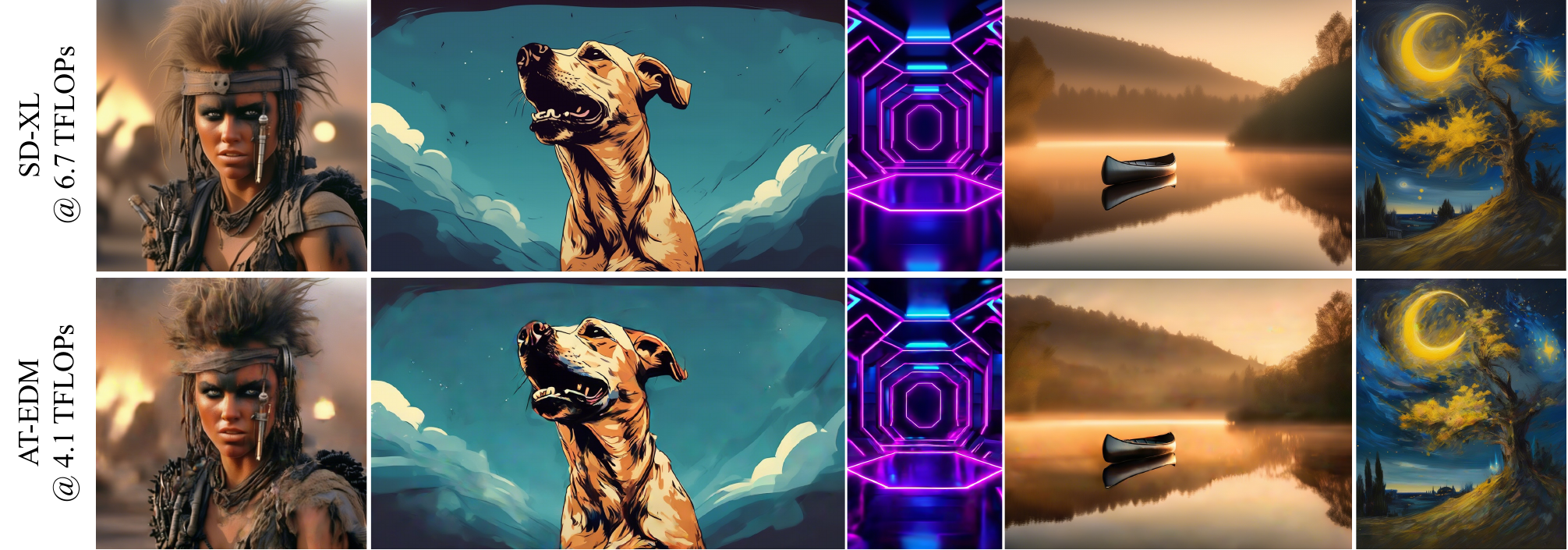}
   \vspace{-1.5em}
   \caption{Examples of applying AT-EDM to SD-XL~\cite{sdxl}. 
   Compared to the full-size model (\textbf{top row}), our accelerated model (\textbf{bottom row}) has around 40\% FLOPs reduction while enjoying competitive generation quality at various aspect ratios.}
   \label{fig:teaser}
   \vspace{-1em}
\end{figure*}

% Diffusion Models (DMs) \cite{dm2015,ddpm} have emerged as powerful tools for text-to-image generation tasks. Compared with Variational Auto-Encoders (VAE) \cite{vae14,vae16}, diffusion models are able to create more photo-realistic images with high fidelity. Although Generative Adversarial Networks (GANs) \cite{stylegan,gigagan} can also generate images with high quality after being trained appropriately with a suitable dataset, diffusion models usually have better generalization ability and can generate more variant images with the same prompt \cite{dmbeatgan}.
Diffusion Models (DMs)~\cite{dm2015,ddpm} have revolutionized computer vision research by achieving state-of-the-art performance in various text-guided content generation tasks, 
including image generation~\cite{ldm}, 
image editing~\cite{imagic}, super resolution~\cite{srdiff}, 3D objects generation~\cite{dreamfusion}, and video generation~\cite{imagenvideo}.
Nonetheless, the superior performance of DMs comes at the cost of an enormous computation budget. 
Although Latent Diffusion Models (LDMs)~\cite{vahdat2021score, ldm} make text-to-image generation much more practical and affordable for normal users, 
their inference process is still too slow. 
For example, on the current flagship mobile phone, generating a single 512px image requires 90 seconds~\cite{snapfusion}.

To address this issue, numerous approaches geared at efficient DMs have been introduced, which can be roughly categorized into two regimes: (1) efficient sampling strategy~\cite{ddim, 4stepldm} and (2) efficient model architecture~\cite{yang2023diffusion, snapfusion}.
While efficient sampling methods can reduce the number of denoising steps, 
they do not reduce the memory footprint and compute cost for each step, making it still challenging to use on devices with limited computational resources. 
On the contrary, an efficient architecture reduces the cost of each step and can be further combined with sampling strategies to achieve even better efficiency. However, most prior efficient architecture works \textbf{require retraining} of the DM backbone, 
which can take thousands of A100 GPU hours. 
% Personal users may not be able to purchase expensive GPUs and not be willing to use cloud computing services due to privacy concerns. 
Moreover, due to different deployment settings on various platforms,
different compression ratios of the backbone model are required, which necessitate multiple retraining runs later. 
Such retraining costs are a big concern even for large companies in the industry.

% For example, DDIM \cite{ddim} changes the forward process to a non-Markovian process and then reduces the number of denoising steps to 50.
% A recent work \cite{4stepldm} reduces the number of denoising steps to 1-4 steps through distillation. 
% (2) Simplification of the U-Net architecture used in each denoising step. BK-SDM \cite{archicomp} obtained a 30\% reduction in floating-point operations (FLOPs) and latency after removing several ResNet and attention blocks in the U-Net through distillation. 

% In the latter part, introduce our methods! 

To this end, we propose the \textbf{Attention-driven Training-free Efficient Diffusion Model (AT-EDM)} framework, 
which accelerates DM inference at run-time without any retraining. 
To the best of our knowledge, training-free architectural compression of DMs is a highly uncharted area. Only one prior work, Token Merging (ToMe)~\cite{tomesd}, addresses this problem. 
While ToMe demonstrates good performance on Vision Transformer (ViT) acceleration \cite{tome}, its performance on DMs still has room to improve.
% Compared to ToME, our approach shows a clear improvement by generating clearer objects with sharper details and better text-image alignment under the same acceleration ratio.
To further enrich research on training-free DM acceleration, we start our study by profiling the floating-point operations (FLOPs) of the state-of-the-art model, Stable Diffusion XL (SD-XL)~\cite{sdxl}, 
through which we find that attention blocks are the dominant workload.
In a single denoising step, we thus propose to dynamically prune redundant tokens to accelerate attention blocks.
We pioneer a fast graph-based algorithm, Generalized Weighted Page Rank (G-WPR), inspired by Zero-TPrune \cite{wpr}, and deploy it on attention maps in DMs to identify superfluous tokens.
Since SD-XL contains ResNet blocks, which require a full number of tokens for the convolution operations, 
we propose a novel similarity-based token copy approach to recover pruned tokens, 
again leveraging the rich information provided by the attention maps.
This token recovery method is critical to maintaining image quality. We find that naive interpolation or padding of pruned tokens adversely impacts generation quality severely. 
In addition to single-step token pruning, we also investigate cross-step redundancy in the denoising process by analyzing the variance of attention maps.
This leads us to a novel pruning schedule, dubbed as Denoising-Steps-Aware Pruning (DSAP), 
in which we adjust the pruning ratios across different denoising timesteps. 
% , inspired by our investigation of the variance of attention maps across steps.
We find DSAP not only significantly improves our method,  
but also helps improve other run-time pruning methods like ToMe \cite{tomesd}. 
Compared to ToMe, our approach shows a clear improvement by generating clearer objects with sharper details and better text-image alignment under the same acceleration ratio.
In summary, our contributions are four-fold:
% AT-EDM is \textit{differentiable}, thus supporting fine-tuning for further performance improvement if there are adequate available computational resources.
% It is \textit{orthogonal} to the previously mentioned efficient sampling and architectural compression works and can thus be used in conjunction with them. We summarize our contributions as follows:

\vspace{-0.2em}
\begin{itemize}
    % \item We profile the FLOPs, parameters, and latency of Stable Diffusion Models (SDMs) in detail, revealing that attention blocks dominate SDM computation. This validates our design choice of focusing on reducing the computational cost of attention blocks.

    \item We propose the AT-EDM framework, 
    which leverages rich information from attention maps to accelerate pre-trained DMs without retraining. 

    \item We design a token pruning algorithm for a single denoising step. We pioneer a fast graph-based algorithm, G-WPR, to identify redundant tokens, and a novel similarity-based copy method to recover missing tokens for convolution. 
    % \yuchen{add G-WPR here?}

    % \item To make the pruned feature maps compatible with convolution operations in ResNet blocks, it is necessary to fill the pruned tokens before passing feature maps to the ResNet blocks. We describe a similarity-based copy method to recover information lost due to pruning.

    \item Inspired by the variance trend of attention maps across denoising steps, we develop the DSAP schedule, which improves generation quality by a clear margin. 
    The schedule also provides improvements over other run-time acceleration approaches, demonstrating its wide applicability.
    
    % \item We use visual examples with manually designed prompts to enable straightforward image quality comparisons. We also evaluate our method on the MS-COCO dataset to obtain FID-CLIP score curves and provide comparisons with state-of-the-art methods. Experimental results demonstrate that with a similar FLOPs budget, our method outperforms state-of-the-art training-free efficiency enhancement methods for SDMs in image quality and text-image alignment.
    \item We use AT-EDM to accelerate a top-tier DM, SD-XL, and conduct both qualitative and quantitative evaluations.
    Noticeably, our method shows comparable performance with an FID score of  28.0 with 40\% FLOPs reduction relative to the full-size SD-XL (FID 27.3), achieving state-of-the-art results. Visual examples are shown in Fig.~\ref{fig:teaser}.

\end{itemize}
\vspace{-0.2em}

% This article is organized as follows. We discuss previous works related to DMs and their efficiency enhancement methods in Section \ref{sec:related}. We describe the proposed AT-EDM framework in detail in Section \ref{sec:methodology}. We present visual and quantitative experimental results in Section \ref{sec:experiments}. Finally, we provide a summary and suggest future work in Section \ref{sec:conclusion}.

% \vspace{-0.5em}
\section{Related Work}
\label{sec:related}
% \vspace{-0.5em}

% In this section, we first introduce the mechanism of our backbone model, Stable Diffusion. Then we introduce existing methods to enhance the efficiency of SDMs, including efficient sampling and architectural compression of the U-Net. Then we introduce a trial of training-free efficiency enhancement for diffusion models in detail, which is our crucial baseline. Finally we introduce the inspritation of our method design.

\paragraph{Text-to-Image Diffusion Models.} 
% DMs gradually add noise to real images and convert them to a normal distribution. This is the diffusion process. Then, 
DMs learn to reverse the diffusion process by denoising samples from a normal distribution step by step. In this manner, the diffusion-based generative models enable high-fidelity image synthesis with variant text prompts \cite{ddpm, dmbeatgan}. However, DMs in the pixel space suffer from large generation latency, which severely limits their applications \cite{dmandgan}. 
% The Cascaded Diffusion Model (CDM) \cite{cascaded} enables application of DMs to cascaded pixel spaces and mitigates this problem. 
The LDM \cite{ldm} was the first to train a Variational Auto-Encoder (VAE) to encode the pixel space into a latent space and apply the DM to the latent space. This reduces computational cost significantly while maintaining generation quality, thus greatly enhancing the application of DMs. Subsequently, several improved versions of the LDM, called Stable Diffusion Models (SDMs), have been released.
% \footnote{https://github.com/Stability-AI/stablediffusion}. 
The most recent and powerful open-source version is SD-XL \cite{sdxl}, which outperforms previous versions by a large margin. SD-XL is our default backbone in this work.
% \yuchen{This part shall be shortened}
% , and please give full name of CDM, LDM, etc.}

\noindent\textbf{Efficient Diffusion Models.} 
%Although an LDM reduces image generation latency by a large margin, it is still much slower than other previous generative models, such as VAE and GAN \cite{gigagan}.
Researchers have made enormous efforts to make DMs more efficient.
% by focusing on sampling in the latent space and attention blocks in these sampling steps. 
%This is a reasonable approach because sampling cost dominates generation cost and the cost of attention layers dominates sampling cost (see our experimental results in Appendix \ref{app:profile}). 
Existing efficient DMs can be divided into two types: 

\noindent
(1) \textbf{Efficient sampling} to reduce the required number of denoising steps \cite{ddim, score, consistency, liu2022pseudo}. A recent efficient sampling work \cite{4stepldm} managed to reduce the number of denoising steps to as low as one. It achieves this by iterative distillation, halving the number of denoising steps each time. 

\noindent
(2) \textbf{Architectural compression} to make each sampling step more efficient \cite{cascaded, dmandgan, snapfusion, yang2023diffusion}. A recent work \cite{archicomp} removes multiple ResNet and attention blocks in the U-Net through distillation. Although these methods can save computational costs while maintaining decent image quality, they require \textit{retraining} of the DM backbone to enhance efficiency, needing thousands of A100 GPU hours. Thus, a training-free method to enhance the efficiency of DMs is needed. Note that our proposed training-free framework, AT-EDM, is \textbf{orthogonal} to these efficiency enhancement methods and can be stacked with them to further improve their efficiency. We provide corresponding experimental evidence in Supplementary Material.

\noindent\textbf{Training-Free Efficiency Enhancement.} Training-free (i.e., post-training) efficiency enhancement schemes have been widely explored for CNNs \cite{tfcnn1,tfcnn2,tfcnn3} and ViTs \cite{tfvit1, ats, tome, wpr}. However, training-free schemes for DMs are still poorly explored. To the best of our knowledge, the only prior work in this field is ToMe \cite{tomesd}. It uses token embedding vectors to obtain pair-wise similarity and merges similar tokens to reduce computational overheads.
% It uses token embedding vectors and merges similar tokens before the self-attention, cross-attention, and feed-forward network to reduce computational overheads. 
% When applied to SD-v1.x and SD-v2.x, ToMe speeds up image generation by up to 50\% (if the resolution is high enough), with only minor quality decay. 
% It has been integrated into Diffusers of Hugging Face. 
% However, we find that ToME does not help much when applied to the state-of-the-art DM backbone, SD-XL, whilst our method achieves a clear improvement over it 
While ToMe achieves a decent speed-up when applied to SD-v1.x and SD-v2.x, we find that it does not help much when applied to the state-of-the-art DM backbone, SD-XL, whilst our method achieves a clear improvement over it (see experimental results in Section \ref{sec:experiments}). 
% (1) architectural change (see supp); (2) better algorithm
This is mainly due to (1) the significant architectural change of SD-XL (see Supplementary Material); (2) our better algorithm design to identify redundant tokens.
% The difference between the performance of ToMe on SD-XL and that on previous versions of SDMs is due to the significant architectural changes in SD-XL: (1) it removes attention blocks at the highest feature level (i.e., the level with most tokens); (2) attention blocks include more than one attention layer (an attention layer is composed of self-attention, cross-attention, and feed-forward network). They prevent the default setting of ToMe from realizing obvious FLOPs savings. To meet the FLOPs budget, we expand the application range of token merging (1) from attention layers at the highest feature level to all attention layers; (2) from self-attention to self-attention, cross-attention, and feed-forward network, when comparing ToMe with AT-EDM.

\noindent\textbf{Exploiting Attention Maps.} We aim to design a method that exploits information present in pre-trained models. ToMe only uses embedding vectors of tokens and ignores the correlation between tokens. We take inspiration from recent image editing works \cite{prompt2prompt, patashnik2023localizing, cao2023masactrl, epstein2023diffusion}, in which attention maps clearly demonstrate which parts of a generated image are more important. 
% For instance, in the generation with the prompt, ``a furry bear watching a bird,” the bear and bird tokens are clearly highlighted in the attention maps. This corresponds to the fact that they constitute key parts of the generated image. 
This inspires us to use the correlations and couplings between tokens indicated by attention maps to identify unimportant tokens and prune them. Specifically, we can convert attention maps to directed graphs, where nodes represent tokens, without information loss. Based on this idea, we develop the G-WPR algorithm for token pruning in a single denoising step.

\noindent\textbf{Non-Uniform Denoising Steps.} Various existing works \cite{fang2023structural, li2023autodiffusion, liu2023oms, yang2023denoising} demonstrate that denoising steps contribute differently to the quality of generated images; thus, it is not optimum to use uniform denoising steps. OMS-DPM \cite{liu2023oms} builds a model zoo and uses different models in different denoising steps. It trains a performance predictor to assist in searching for the optimal model schedule. DDSM \cite{yang2023denoising} employs a spectrum of neural networks and adapts their sizes to the importance of each denoising step. AutoDiffusion \cite{li2023autodiffusion} employs evolutionary search to skip some denoising steps and some blocks in the U-Net. Diff-Pruning \cite{fang2023structural} uses a Taylor expansion over pruned timesteps to disregard non-contributory diffusion steps. All existing methods either require an intensive training/fine-tuning/searching process to obtain and deploy the desired denoising schedule or are not compatible with our proposed G-WPR token pruning algorithm due to the U-Net architecture change. On the contrary, based on our investigation of the variance of attention maps across denoising steps, we propose DSAP. Its schedule can be determined via simple ablation experiments and it is compatible with any token pruning scheme. DSAP can potentially be migrated to existing efficient DMs to help improve their image quality.
\section{Methodology}
\label{sec:methodology}
% \vspace{-0.5em}

\begin{figure}[t]
  \centering
  % \vspace{-1em}
   \includegraphics[width=\linewidth]{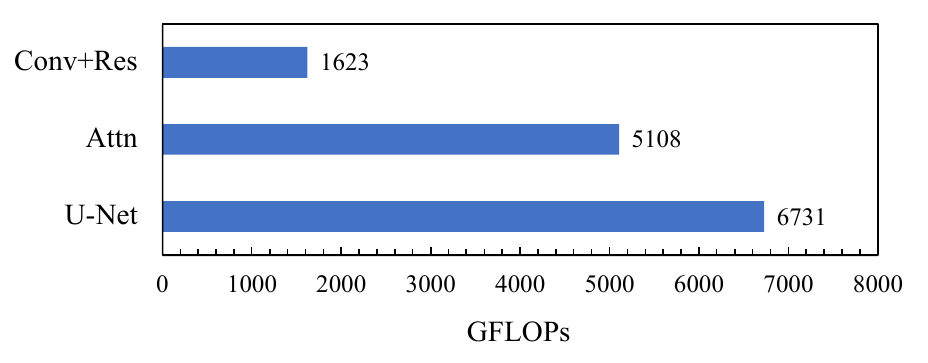}
   \vspace{-2.5em}
   \caption{U-Net FLOPs breakdown of SD-XL~\cite{sdxl} measured with 1024px image generation. Among components of U-Net (convolution blocks, ResNet blocks, and attention blocks), attention blocks cost the most.
   %Attention blocks cost the most.
   % ``Conv+Res" represents the FLOPs of convolution and ResNet blocks, ``Attn" represents the FLOPs of attention blocks. 
   }
   \vspace{-1.5em}
   \label{fig:breakdown}
\end{figure}

We start our investigation by profiling the FLOPs of the state-of-the-art DM, SD-XL, as shown in Fig.~\ref{fig:breakdown}. Noticeably, among compositions of the sampling module (U-Net), attention blocks, which consist of several consecutive attention layers, dominate the workload for image generation. Therefore, we propose AT-EDM to accelerate attention blocks in the model through token pruning. AT-EDM contains two important parts: a single-denoising-step token pruning scheme and the DSAP schedule. We provide an overview of these two parts and then discuss them in detail.

\subsection{Overview}

% Inspired by semantically meaningful attention maps, we propose our AT-EDM framework, as shown in 
Fig.~\ref{fig:overview} illustrates the two main parts of AT-EDM: 

\begin{figure*}[htbp]
  \centering
  % \vspace{-2em}
   \includegraphics[width=0.95\linewidth]{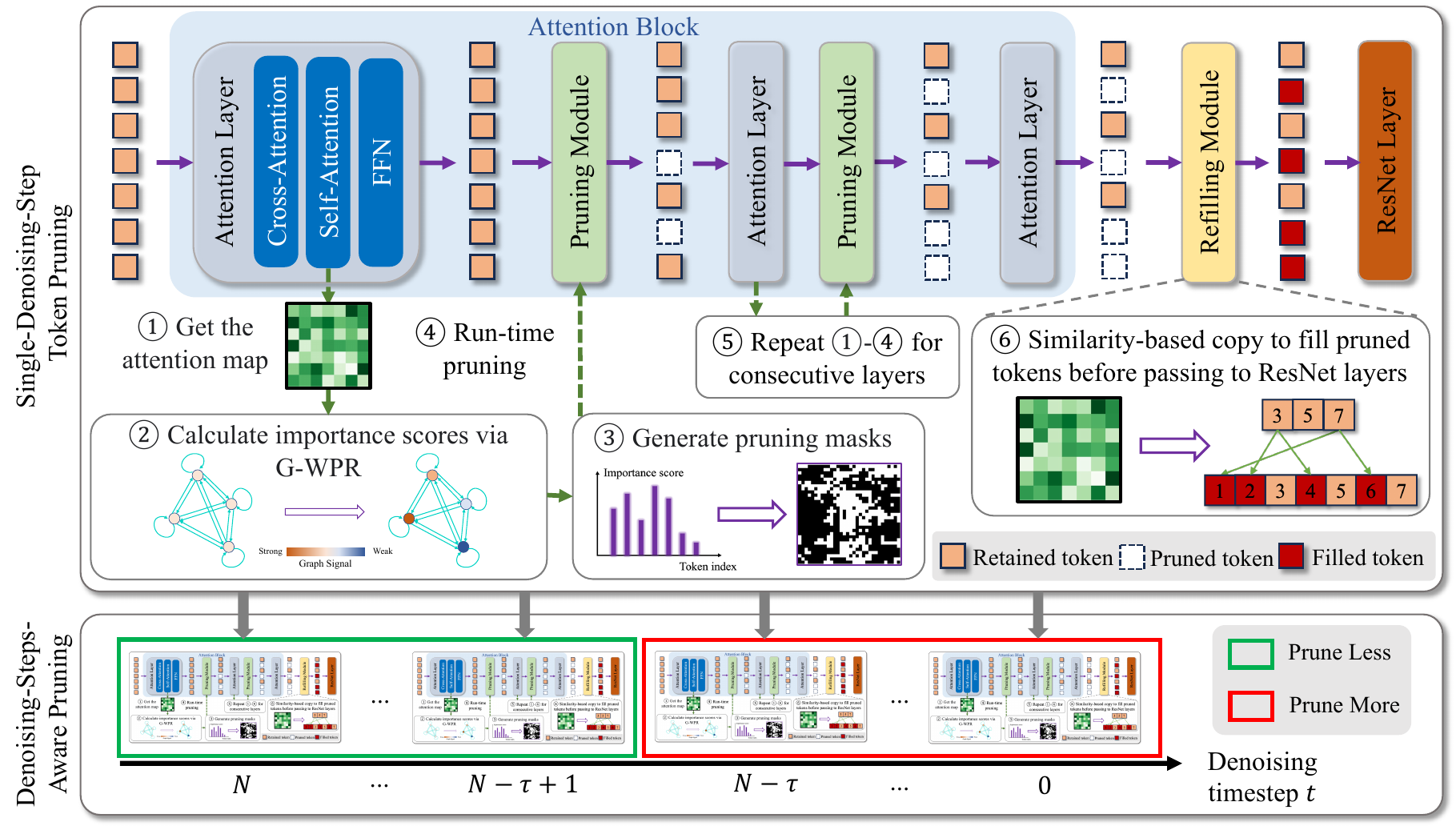}
   \vspace{-1em}
   \caption{Overview of our proposed efficiency enhancement framework \textbf{AT-EDM}. \textbf{Single-Denoising-Step Token Pruning:} (1) We get the attention map from self-attention. (2) We calculate the importance score for each token using G-WPR. (3) We generate pruning masks. (4) We apply the masks to tokens after the feed-forward network to realize token pruning. (5) We repeat Steps (1)-(4) for each consecutive attention layer. (6) Before passing feature maps to the ResNet block, we recover pruned tokens through similarity-based copy. 
   \textbf{Denoising-Steps-Aware Pruning Schedule:}  In early steps, we propose to prune fewer tokens and to have less FLOPs reduction. In later steps, we prune more aggressively for higher speedup.
   %This pruning schedule is referred to as ``DSAP" in later text.
   }
   \vspace{-1em}
   \label{fig:overview}
\end{figure*}

\vspace{-1em}

\noindent\paragraph{Part I: Token pruning scheme in a single denoising step.} 

\noindent
\textbf{Step 1:} We obtain the attention maps from an attention layer in the U-Net. We can potentially obtain the attention maps from self-attention or cross-attention. We compare the two choices and analyze them in detail through ablation experiments.

\noindent
\textbf{Step 2:} We use a scoring module to assign an importance score to each token based on the obtained attention map. We use an algorithm called G-WPR to assign importance scores to each token. This is described in Section \ref{sec:tpsingle}. 

\noindent
\textbf{Step 3:} We generate pruning masks based on the calculated importance score distribution. Currently, we simply use the top-$k$ approach to determine the retained tokens, i.e., prune tokens with less importance scores. 
% To make the pruning schedule dependent on the difficulty of the generation task, it is important to use special techniques, such as inverse transform sampling \cite{ats}.

\noindent
\textbf{Step 4:} We use the generated mask to perform token pruning. We do this after the feed-forward layer of attention layers. We may also perform pruning early before the feed-forward layers. 
%However, this may adversely impact the quality of generated images. 
We provide ablative experimental results for it in Supplementary Material.

\noindent
\textbf{Step 5:} We repeat Steps 1-4 for each consecutive attention layer. Note that we do not apply pruning to the last attention layer before the ResNet layer.
%Instead of repeating the whole process, we may also reuse the obtained attention maps across consecutive attention blocks. The necessity of doing so is discussed later.

\noindent
\textbf{Step 6:} Finally, before passing the pruned feature map to the ResNet block, we need to fill (i.e., try to recover) the pruned tokens. A simple approach is to pad zeros, which means we do not fill anything. The method that we currently use is to copy tokens to corresponding locations based on similarity. This is described in detail in Section \ref{sec:tpsingle}.

\vspace{-1em} 

\noindent\paragraph{Part II: DSAP schedule.} Attention maps in early denoising steps are more chaotic and less informative than those in later steps, which is indicated by their low variance. Thus, they have a weaker ability to differentiate unimportant tokens \cite{prompt2prompt}. Based on this intuition, we design the DSAP schedule that prunes fewer tokens in early denoising steps. Specifically, we select some attention blocks in the up-sampling and down-sampling stages and leave them unpruned, since they contribute more to the generated image quality than other attention blocks \cite{snapfusion}. We demonstrate the schedule in detail in Section \ref{sec:hetero}.

% \vspace{0.5em}

% Our method is completely training-free. Thus, it can be applied to any pre-trained DM without the need for cumbersome fine-tuning and thus obtain off-the-shelf computational savings. We use attention maps to prune less important tokens and recover them based on similar retained tokens, ensuring high-fidelity image generation with token pruning.

% Overall, our method can be divided into three modes: (1) frozen backbone and parameter-free scoring module; (2) frozen backbone and scoring module with learnable parameter; (3) fine-tuned backbone and scoring module with learnable parameter. Starting from frozen backbone and parameter-free scoring module, we may gradually tune the scoring module and the backbone. From the first mode to the third mode, we trade universality for efficiency and image quality. Currently we are still focusing on the left mode. 

% \begin{figure}[t]
%   \centering
%    \includegraphics[width=\linewidth]{sec/modes.png}

%    \caption{The three modes of our proposed efficiency enhancement framework, Edmea.}
%    \label{fig:modes}
% \end{figure}

\subsection{Part I: Token Pruning in a Single Step}
\label{sec:tpsingle}

% Next, we first introduce the used notations and then describe our G-WPR algorithm and the similarity-based copy method to recover pruned tokens.

% \subsubsection{Notation}

\noindent\textbf{Notation.} Suppose $\mathbf{A}^{(h,l)}\in \mathbb{R}^{M \times N}$ is the attention map of the $h$-th head in the $l$-th layer. It reflects the correlations between $M$ Query tokens and $N$ Key tokens. We refer to $\mathbf{A}^{(h,l)}$ as $\mathbf{A}$ for simplicity in the following discussion. Let $A_{i,j}$ denote its element in the $i$-th row, $j$-th column. $\mathbf{A}$ can be thought of as the adjacency matrix of a directed graph in the G-WPR algorithm. In this graph, the set of nodes with input (output) edges is referred to as $\Phi_{in}$ ($\Phi_{out}$). Nodes in $\Phi_{in}$ ($\Phi_{out}$) represent Key (Query) tokens, i.e., $\Phi_{in}=\{k_j\}_{j=1}^N$ ($\Phi_{out}=\{q_i\}_{i=1}^M$). Let $s_K^t$ ($s_Q^t$) denote the vector that represents the importance score of Key (Query) tokens in the $t$-th iteration of the G-WPR algorithm. In the case of self-attention, Query tokens are the same as Key tokens. Specifically, we let $\{x_i\}^N_{i=1}$ denote the $N$ tokens and $s$ denote their importance scores in the description of our token recovery method.

% \subsubsection{The G-WPR Algorithm}
% \label{sec:wpr}

% Page rank is an algorithm used by Google to rank the importance of web pages based on links between them. Inspired by this, the WPR algorithm was proposed in \cite{wpr}. 
\noindent\textbf{The G-WPR Algorithm.} WPR \cite{wpr} uses the attention map as an adjacency matrix of a directed complete graph. It uses a graph signal to represent the importance score distribution among nodes in this graph. This signal is initialized uniformly. WPR uses the adjacency matrix as a graph operator, applying it to the graph signal iteratively until convergence. In each iteration, each node votes for which node is more important. The weight of the vote is determined by its importance in the last iteration.
However, WPR, as proposed in \cite{wpr}, constrains the used attention map to be a self-attention map. Based on this, we propose the G-WPR algorithm, which is compatible with both self-attention and cross-attention, as shown in Algorithm \ref{alg:prgraph}. The attention from Query $q_i$ to Key $k_j$ weights the edge from $q_i$ to $k_j$ in the graph generated by $\mathbf{A}$. In each iteration of the vanilla WPR, by multiplying with the attention map, we map the importance of Query tokens $s^t_Q$ to the importance of Key tokens $s^{t+1}_K$, i.e., each node in $\Phi_{out}$ votes for which $\Phi_{in}$ node is more important. For self-attention, $s^{t+1}_Q=s^{t+1}_K$ since Query and Key tokens are the same. For cross-attention, Query tokens are image tokens and Key tokens are text prompt tokens. Based on the intuition that important image tokens should devote a large portion of their attention to important text prompt tokens, we define function $f(\mathbf{A}, s_K)$ that maps $s^{t+1}_K$ to $s^{t+1}_Q$. One entropy-based implementation is 

\vspace{-1em}
% \begin{equation}
% \small
%     s^{t+1}_Q(q_i) = f(\mathbf{A}, s^{t+1}_K)=\frac{\sum_{j=1}^{N}\mathbf{A}(q_{i}, k_{j})\cdot s^{t+1}_K(k_j)}{-\sum_{j=1}^{N}\mathbf{A}(q_{i}, k_{j})\cdot\ln{\mathbf{A}(q_{i}, k_{j})}}
% \end{equation}

\begin{equation}
\small
    s^{t+1}_Q(q_i) = f(\mathbf{A}, s^{t+1}_K)=\frac{\sum_{j=1}^{N}A_{i,j}\cdot s^{t+1}_K(k_j)}{-\sum_{j=1}^{N}A_{i,j}\cdot\ln{A_{i,j}}}
\end{equation}

\noindent
where $A_{i,j}$ is the attention from Query $q_i$ to Key $k_j$. This is the default setting for cross-attention-based WPR in the following sections. We discuss and compare other implementations in Supplementary Material. Note that for self-attention, $f(\mathbf{A}, s^{t+1}_K)=s^{t+1}_K$.
The G-WPR algorithm has an $O(M\times N)$ complexity, where $M$ ($N$) is the number of Query (Key) tokens. We employ this algorithm in each head and then obtain the root mean square of scores from different heads (to reward tokens that obtain very high importance scores in a few heads).

\begin{algorithm}
% \scriptsize
\small
\caption{The G-WPR algorithm for both self-attention and cross-attention}\label{alg:prgraph}
\begin{algorithmic}
\Require $M, N>0$ is the number of nodes in $\Phi_{out}, \Phi_{in}$; $\mathbf{A}\in \mathbb{R}^{M \times N}$; $s_Q\in \mathbb{R}^{M}, s_K\in \mathbb{R}^{N}$; $f(\mathbf{A},s_k)$ maps the importance of Key to that of Query
\Ensure  $s\in \mathbb{R}^{M}$ represents the importance score of image tokens
\State $s^{0}_Q \gets \frac{1}{M} \times \textbf{$e_M$}$ % \Comment{Initialize the graph signal uniformly}
\State $t \gets 0$
\While{$(|s^{t}_Q - s^{t-1}_Q|> \epsilon) ~\textbf{or}~(t=0)$} % \Comment{Continue iterating if not converged}
\State $s^{t+1}_K \gets \mathbf{A}^T \times s^{t}_Q$  % \Comment{Use the adjacency matrix as a graph shift operator}
\State $s^{t+1}_Q \gets f(\mathbf{A},s^{t+1}_K) $ % \Comment{Map the importance of Keys to that of Querys}
\State $s^{t+1}_Q \gets s^{t+1}_Q / |s^{t+1}_Q|$
\State $t \gets t+1$
\EndWhile
\State $s \gets s^{t}_Q$
\end{algorithmic}
\end{algorithm}

\begin{figure}[t]
  \centering
   \includegraphics[width=\linewidth]{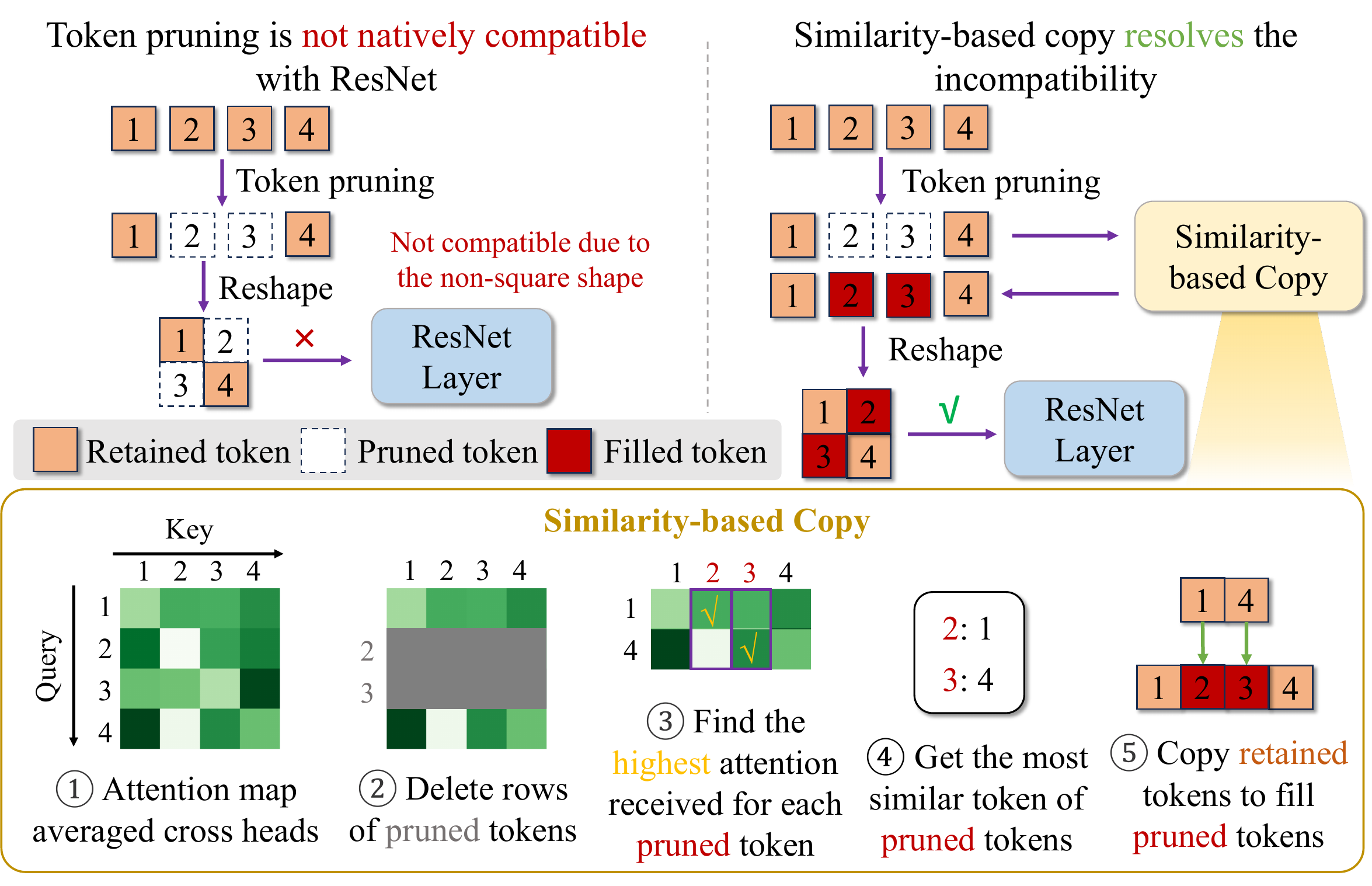}
   \vspace{-1.5em}
   \caption{Our similarity-based copy method for token recovering resolves the incompatibility between token pruning and ResNet. Token pruning incurs the non-square shape of feature maps and thus is not compatible with ResNet. To address this issue, we propose similarity-based copy to recover the pruned tokens. It first averages the attention map across heads and deletes the rows of pruned tokens to avoid selecting them as the most similar one. Then, it finds the source of the highest attention received for each pruned token and copies the corresponding retained tokens for recovery. After recovering, the tokens can be translated into a spatially-complete feature map to serve as input to ResNet blocks.}
   \vspace{-1.5em}
   \label{fig:sim_copy}
\end{figure}

\noindent\textbf{Recovering Pruned Tokens.} We have fewer tokens after token pruning, leading to efficiency enhancement. However, retained tokens form irregular maps and thus cannot be used for convolution, as shown in Fig.~\ref{fig:sim_copy}. We need to recover the pruned tokens to make them compatible with the following convolutional operations in the ResNet layer.
% (I) Padding Zeros. (II) Interpolation. (III) Direct copy. (check supp for details)

\noindent
\textbf{(I) Padding Zeros.} One straightforward way to do this is to pad zeros.
However, to maintain the high quality of generated images, we hope to recover the pruned tokens as precisely as possible, as if they were not pruned. 

\noindent
\textbf{(II) Interpolation.} Interpolation methods, such as bicubic interpolation, are not suitable in this context. To use the interpolation algorithm, we first pad zeros to fill the pruned tokens and form a feature map of size $N\times N$. Then we downsample it to $\frac{N}{2}\times \frac{N}{2}$ and upsample it back to $N\times N$ with the interpolation algorithm. We keep the values of retained tokens fixed and only use the interpolated values of pruned tokens. Due to the high pruning rates (usually larger than 50\%), most tokens that represent the background get pruned, leading to lots of pruned tokens that are surrounded by other pruned tokens instead of retained tokens. Interpolation algorithms assign nearly zero values to these tokens. 

\noindent
\textbf{(III) Direct copy.} Another possible method is to use the corresponding values before pruning is applied (i.e., before being processed by the following attention layers) to fill the pruned tokens. The problem with this method is that the value distribution changes significantly after being processed by multiple attention layers, and copied values are far from the values of these tokens if they are not pruned and are processed by the following attention layers.

To avoid the effect of distribution shift, we propose the \textbf{similarity-based copy} technique, as shown in Fig.~\ref{fig:sim_copy}. Instead of copying values that are not processed by attention layers, we select tokens that are similar to pruned tokens from the retained tokens. We use the self-attention map to determine the source of the highest attention received for each pruned token and use that as the most similar one. This is based on the intuition that attention from token $x_a$ to token $x_b$, $A_{a,b}$, is determined by two factors: (1) importance of token $x_b$, i.e., $s(x_b)$, and (2) similarity between token $x_a$ and $x_b$. If we observe the attention that $x_b$ receives, i.e., compare $\{A_{i,b}\}_{i\in N}$, since $s(x_b)$ is fixed, index $i=\eta$ that maximizes $\{A_{i,b}\}_{i\in N}$ is the index of the most similar token, i.e., $x_{\eta}$. Finally, we copy the value of token $x_{\eta}$ to fill (i.e., recover) the pruned token $x_b$.

\subsection{Part II: Denoising-Steps-Aware Pruning}

\label{sec:hetero}

\begin{figure}[t]
  \centering
  % \vspace{-1em}
  % \fbox{\rule{0pt}{0.5in} \rule{0.9\linewidth}{0pt}}
  \includegraphics[width=\linewidth]{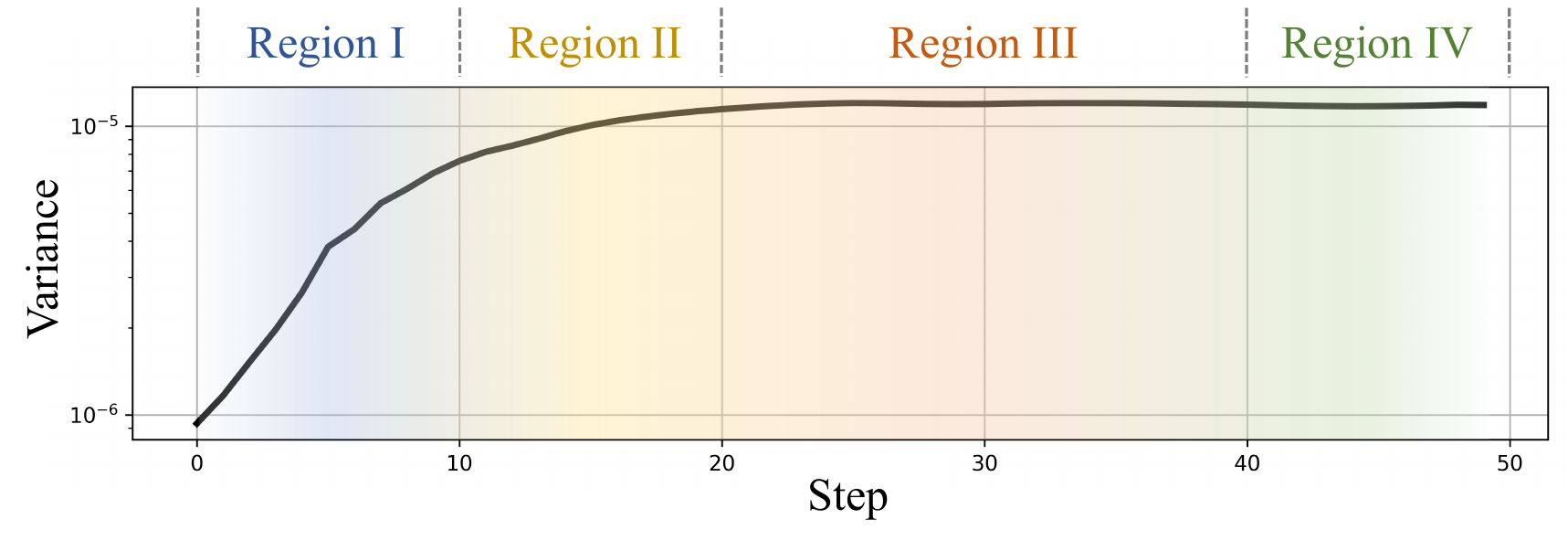}
  % \vspace{-2.5em}
   \caption{Variance of attention maps in different denoising steps. We divide the denoising steps into four typical regions: (I) Very-early steps: Variance of attention maps is small and increases rapidly. (II) Mid-early steps: Variance of attention maps is large and increases slowly. (III) Middle steps: Variance of attention maps is large and almost constant. (IV) Last several steps.}
   \vspace{-1.0em}
   \label{fig:var_attn}
\end{figure}

Early denoising steps determine the layout of generated images and, thus, are crucial. On the contrary, late denoising steps aim at refining the generated image, natively including redundant computations since many regions of the image do not need refinement.
In addition, \textit{early denoising steps have a weaker ability to differentiate unimportant tokens}, and late denoising steps yield informative attention maps and differentiate unimportant tokens better. To support this claim, we investigate the variance of feature maps in different denoising steps, as shown in Fig. \ref{fig:var_attn}. It indicates that attention maps in early steps are more uniform. They assign similar attention scores to both important and unimportant tokens, making it harder to precisely identify unimportant tokens and prune them in early steps. Based on these intuitions, we propose DSAP that employs a \textbf{prune-less schedule} in early denoising steps by leaving some of the layers unpruned. 
% We explore setting the boundary between prune-less and normal schedule in different regions shown in Fig.~\ref{fig:var_attn} (Section \ref{sec:ablation}). To further consolidate these intuitions, we also investigate a more aggressive pruning schedule in early denoising steps and find it is inferior to our current approach (see Supplementary Materials).

\noindent\textbf{The Prune-Less Schedule.} In SD-XL, each down-stage includes two attention blocks and each up-stage includes three attention blocks (except for stages without attention). The mid-stage also includes one attention block. Each attention block includes 2-10 attention layers. In our prune-less schedule, we select some attention blocks to not perform token pruning. Since previous works \cite{snapfusion, archicomp} indicate that the mid-stage contributes much less to the generated image quality than the up-stages and down-stages, we do not select the attention block in the mid-stage. Based on the ablation study, we choose to leave the first attention block in each down-stage and the last attention block in each up-stage unpruned. We use this prune-less schedule for the first $\tau$ denoising steps. We explore setting $\tau$ in different regions shown in Fig.~\ref{fig:var_attn} and find $\tau=15$ is the optimal choice. We present all the related ablative experimental results in Section \ref{sec:ablation}. A detailed description of the less aggressive pruning schedule is provided in Supplementary Material. To further consolidate our intuitions, we also investigate a more aggressive pruning schedule in early denoising steps and find it is inferior to our current approach (see Supplementary Material).

\section{Experimental Results}
\label{sec:experiments}

\begin{figure*}[htbp]
  \centering
  % \vspace{-2em}
   \includegraphics[width=0.9\linewidth]{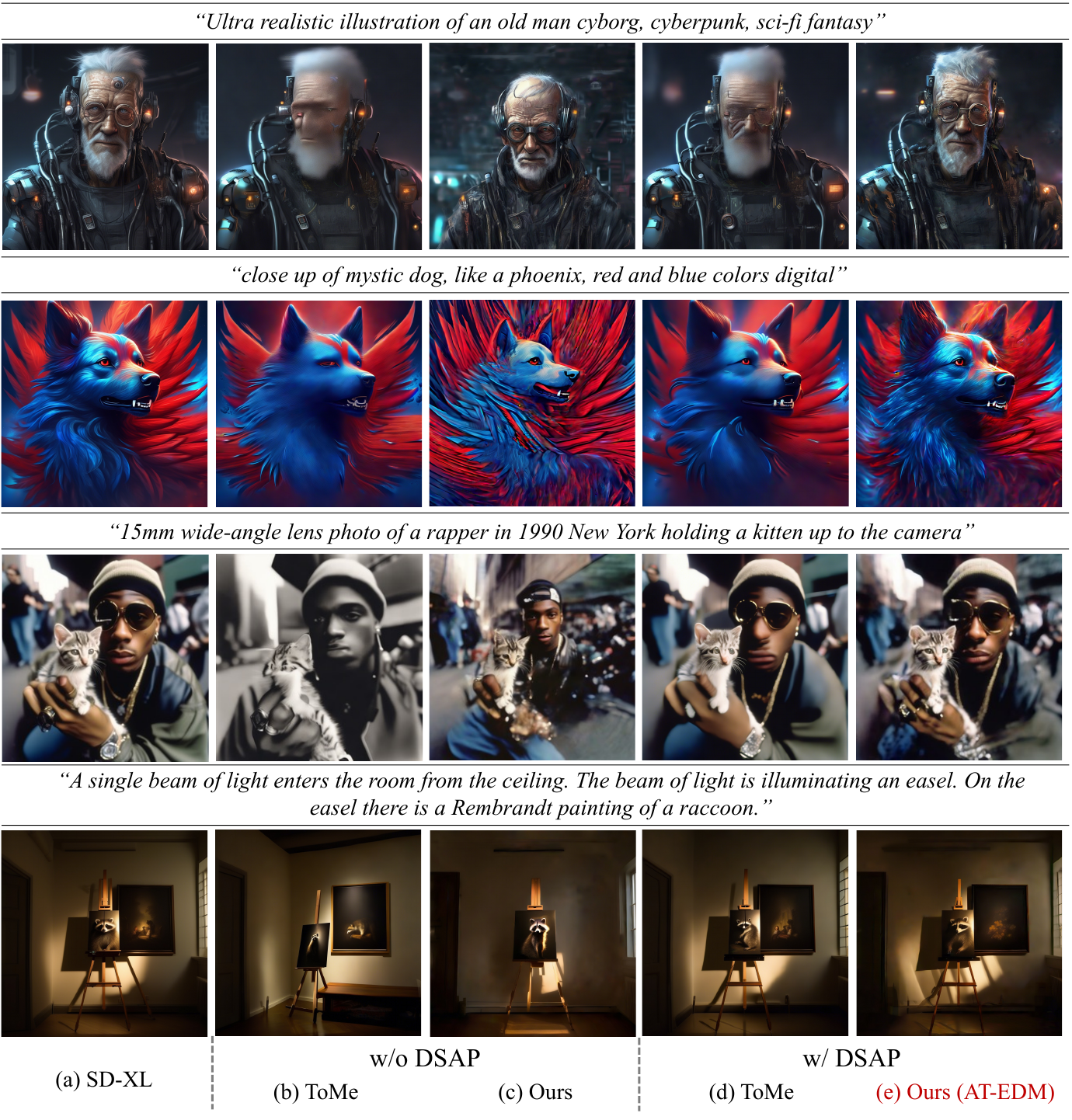}
   \vspace{-1.25em}
   \caption{Comparing AT-EDM to the state-of-the-art approach, ToMe~\cite{tome}. While the full-size SD-XL~\cite{sdxl} (\textbf{Col. a}) consumes 6.7 TFLOPs, we compare the accelerated models (\textbf{Col.~b-e}) at the same budget of 4.1 TFLOPs.  
   Compared to ToMe, we find that AT-EDM's token pruning algorithm provides clearer generated objects with sharper details and finer textures, 
   and a better text-image alignment where it better retains the semantics in the prompt (see the fourth row). 
   Moreover, we find that DSAP provides better structural layout of the generated images, which is effective for both ToMe and our approach.
   AT-EDM combines the novel token pruning algorithm and the DSAP schedule (\textbf{Col.~e}), outperforming the state of the art.}
       \vspace{-1.25em}
   \label{fig:visualization}
\end{figure*}

In this section, we evaluate AT-EDM and ToMe on SD-XL. We provide both visual and quantitative experimental results to demonstrate the advantages of AT-EDM over ToMe.

\subsection{Experimental Setup}

\noindent\textbf{Common Settings.} We implement both our AT-EDM method and ToMe on the official repository
% \footnote{https://github.com/Stability-AI/generative-models} 
of SD-XL and evaluate their performance. The resolution of generated images is 1024$\times$1024 pixels and the default FLOPs budget for each denoising step is assumed to be 4.1T, which is 38.8\% smaller than that of the original model (6.7T) unless otherwise noted. 
% [We set the batch size to 1 when we measure the FLOPs cost.] 
The default CFG-scale for image generation is 7.0 unless otherwise noted. We set the total number of sampling steps to 50. We use the default sampler of SD-XL, i.e., EulerEDMSampler. 

\noindent\textbf{AT-EDM.} For a concise design, we only insert a pruning layer after the first attention layer of each attention block and set the pruning ratio for that layer to $\rho$. To meet the FLOPs budget of 4.1T, we set $\rho=63\%$. For the DSAP setting, we choose to leave the first attention block in each down-stage and the last attention block in each up-stage unpruned. We use this prune-less schedule for the first $\tau=15$ denoising steps. 
% \textbf{move to supplementary} [We use the \texttt{THOP} library\footnote{https://github.com/Lyken17/pytorch-OpCounter} to measure the FLOPs cost of different DMs. Since the vanilla implementation of THOP cannot correctly measure the FLOPs cost of self-attention in Stable Diffusion Models, we added a calibration block to it (more details in Supplementary Materials). ]

\noindent\textbf{ToMe.} The SD-XL architecture has changed significantly compared to previous versions of SDMs (see Supplementary Material). Thus, the default setting of ToMe does not lead to enough FLOPs savings. To meet the FLOPs budget, it is necessary to use a more aggressive merging setting. Therefore, we expand the application range of token merging (1) from attention layers at the highest feature level to all attention layers, and (2) from self-attention to self-attention, cross-attention, and the feedforward network. We set the merging ratio $r=50\%$ to meet the FLOPs budget of 4.1T. 

\noindent\textbf{Evaluations.} We first compare the generated images with manually designed challenging prompts in Section \ref{sec:visual}. Then, we report FID and CLIP scores of zero-shot image generation on the MS-COCO 2017 validation dataset \cite{mscoco} in Section \ref{sec:fidclip}. Tested models generate 1024$\times$1024 px images based on the captions of 5k images in the validation set. We provide ablative experimental results and analyze them in Section \ref{sec:ablation} to justify our design choices. 
% Note that we obtain all results shown in this section off-the-shelf without any fine-tuning or retraining. 
We provide more implementation details in Supplementary Material. 

\subsection{Visual Examples for Qualitative Analysis}
\label{sec:visual}

We use manually designed challenging prompts to evaluate ToMe and our proposed AT-EDM framework. The generated images are compared in Fig.~\ref{fig:visualization}. We compare more generated images in Supplementary Material. Visual examples indicate that with the same FLOPs budget, AT-EDM demonstrates better \textbf{main object preservation} and \textbf{text-image alignment} than ToMe. For instance, in the first example, AT-EDM preserves the main object, the face of the old man, much better than ToMe does. AT-EDM's strong ability to preserve the main object is also exhibited in the second example. ToMe loses high-frequency features of the main object, such as texture and hair, while AT-EDM retains them well, even without DSAP. 
%On the contrary, ToMe not only generates a chaotic face but also changes the gender of this person. 
The third example again illustrates the advantage of AT-EDM over ToMe in preserving the rapper's face. 
The fourth example uses a relatively complex prompt that describes relationships between multiple objects.
%such as ``rapper'',  ``holding'', ``kitchen'', and so on. AT-EDM is the only pruning method that maintains the hand that holds the cat.
% The image generated by ToMe loses the information that the beam of light enters the room \textit{"from the ceiling."} What's more, it 
ToMe misunderstands \textit{"a Rembrandt painting of a raccoon"} as being a random painting on the easel and a painting of a raccoon on the wall. On the contrary, the image generated by AT-EDM understands and preserves these relationships very well, even without DSAP. 
As a part of our AT-EDM framework, DSAP is not only effective in AT-EDM but also beneficial to ToMe in improving image quality and text-image alignment. When we deploy DSAP in ToMe, we select corresponding attention blocks to not perform token merging, while keeping the FLOPs cost fixed. 

% Besides the face reservation of the Pomeranian in blue circles, our generated image also captured the information of “sitting on the throne”, as shown in the pink circle. On the contrary, in the image generated by ToMe, the Pomeranian sits on the floor. Our generated images also emphasized the tiger characteristics of soldiers, as shown in the red and green circle. 

% Another example is a cute corgi lives in a house made out of sushi. Again, our method has a much better main object reservation, compared with ToMe, as shown in the blue circle. What’s more, our method maintained the main structure of the house and represented the sushi much better than ToMe, as shown in the pink circle.

% The two visual examples in Fig. \ref{fig:tome_vis} also exhibit the image quality decay obviously. As we can see, the face of dogs becomes very chaotic after ToMe is applied. And the face of one tiger soldier, which is in the green circle, nearly disappears.

% \subsection{Comparison in terms of FLOPs}

% \begin{figure}[t]
%   \centering
%    \includegraphics[width=\linewidth]{sec/tome.png}

%    \caption{An attention block with ToMe applied in the U-Net. Figure adapted from [10]}
%    \label{fig:tome}
% \end{figure}

% \begin{figure}[t]
%   \centering
%    \includegraphics[width=\linewidth]{sec/tome_save.png}

%    \caption{The speed-up of ToMe and random dropping in terms of latency. Evaluated on a single A100-40G; generate 1024x1024 px images with batch size equals to 3.}
%    \label{fig:tome_save}
% \end{figure}

\subsection{Quantitative Evaluations}
\label{sec:fidclip}

\begin{figure}[t]
  \centering
   \includegraphics[width=\linewidth]{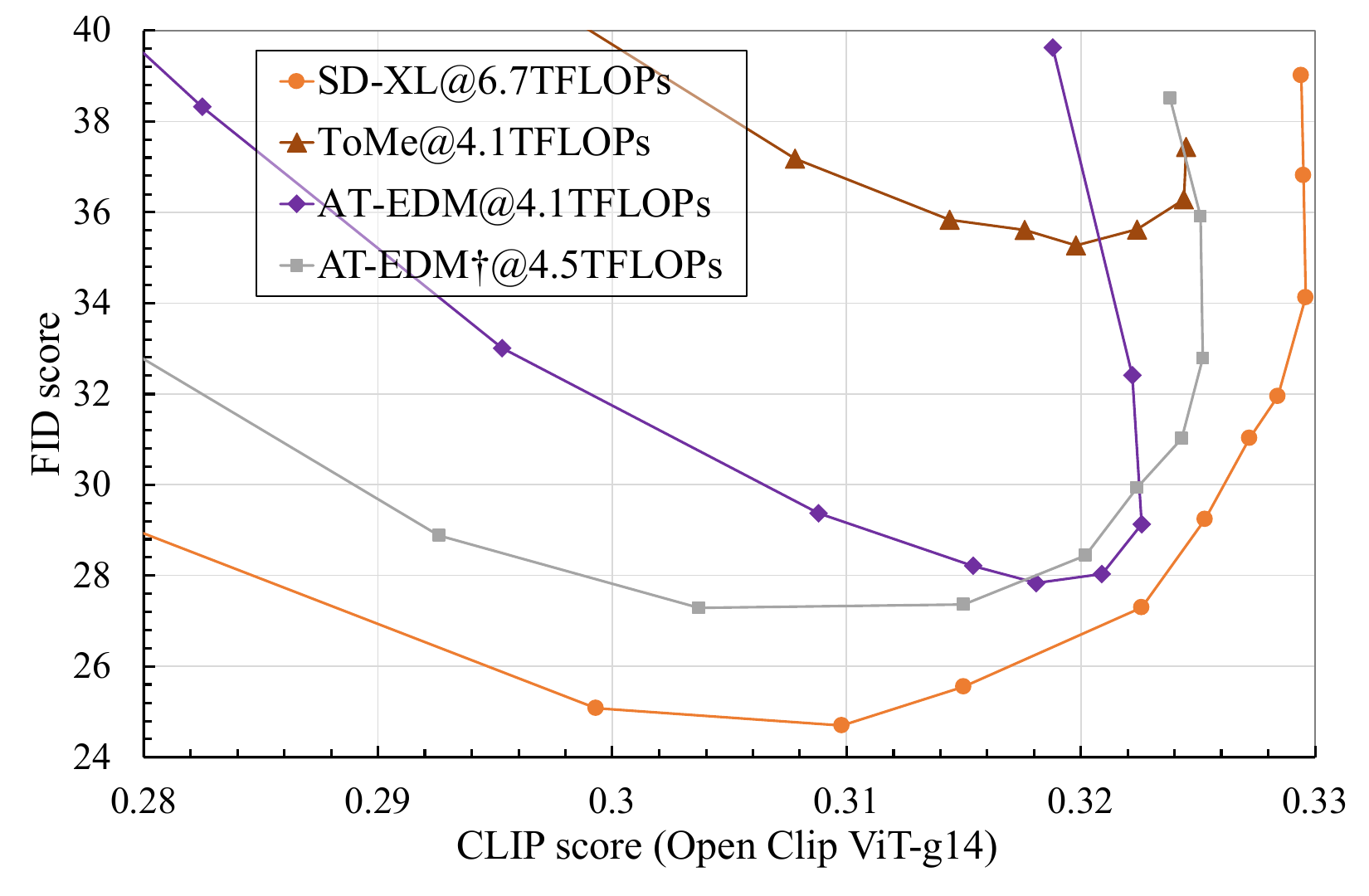}
   \vspace{-2.25em}
   \caption{FID-CLIP score curves. The used CFG scales are [1.0, 1.5, 2.0, 2.5, 3.0, 4.0, 5.0, 6.0, 7.0, 9.0, 12.0, 15.0]. This figure is zoomed in to the bottom-right corner to show the comparison between the best trade-off points. AT-EDM outperforms ToMe by a clear margin. See complete curves in Supplementary Material.}
   \vspace{-1em}
   \label{fig:tome_fid}
\end{figure}

\noindent\textbf{FID-CLIP Curves.} We explore the trade-off between the CLIP and FID scores through various Classifer-Free Guidance (CFG) scales. We show the results in Fig.~\ref{fig:tome_fid}. AT-EDM$^\dagger$ does not deploy pruning at the second feature level (see Supplementary Material). It indicates that for most CFG scales, AT-EDM not only lowers the FID score but also results in higher CLIP scores than ToMe, implying that images generated by AT-EDM not only have better quality but also better text-image alignment. Specifically, when the CFG scale equals 7.0, AT-EDM results in [FID, CLIP] = [28.0, 0.321], which is almost the same as the full-size one ([27.3, 0.323], \texttt{CFG\_scale}=4.0). For comparison, ToMe results in [35.3, 0.320] with a CFG scale of 7.0. Thus, \textbf{AT-EDM reduces the FID gap from 8.0 to 0.7.} 
% We provide the comparison under more FLOPs budgets in Supplementary Materials.

\noindent\textbf{Various FLOPs Budgets.} We deploy ToMe and AT-EDM on SD-XL under various FLOPs budgets and quantitatively compare their performance in Table \ref{tab:flops}. The FLOPs cost in this table refers to the average FLOPs cost of a denoising step. Table \ref{tab:flops} indicates that AT-EDM achieves better image quality than ToMe (lower FID scores) under all FLOPs budgets. When the FLOPs budget is extremely low (less than 50\% of the full model), ToMe achieves higher CLIP scores than AT-EDM. When the FLOPs saving is 30-40\%, AT-EDM achieves not only better image quality (lower FID scores) but also better text-image alignment (higher CLIP scores) than ToMe. Note that under the same CFG-scale, AT-EDM achieves a lower FID score than the full-size model while reducing FLOPs by 32.8\%. In the case that it trades text-image alignment for image quality (via reducing the CFG scale to 4.0), AT-EDM achieves \textbf{not only a lower FID score but also a higher CLIP score than the full-size model while reducing FLOPs by 32.8\%}. We provide more visual examples under various FLOPs budgets in Supplementary Material.

\begin{table}[t]
  \caption{Deploying ToMe and AT-EDM in SD-XL under different FLOPs budgets. We generate all images with the CFG-scale of 7.0, except for SD-XL$^\dagger$, for which we use a CFG-scale of 4.0.}
  % \vspace{-1em}
  \label{tab:flops}
  % \small
  \centering
  \begin{tabular}{cccc}
    \toprule
    %\multicolumn{2}{c}{Part}                   \\
    %\cmidrule(r){1-2}
    Model    &  FID     & CLIP & TFLOPs \\
    \midrule
    SD-XL & 31.94 & 0.3284 & 6.7     \\
    SD-XL$^\dagger$ & 27.30 & 0.3226 & 6.7 \\
    ToMe-a & 58.76 & 0.2954 & 2.9 \\
    \rowcolor{cyan!30}
    AT-EDM-a & 52.00 & 0.2784 & 2.9 \\
    ToMe-b & 40.94 & 0.3154 & 3.6 \\
    \rowcolor{cyan!30}
    AT-EDM-b & 29.80 & 0.3095 & 3.6 \\
    ToMe-c & 35.27 & 0.3198 & 4.1 \\
    \rowcolor{cyan!30}
    AT-EDM-c & 28.04 & 0.3209 & 4.1 \\
    ToMe-d & 32.46 & 0.3235 & 4.6 \\
    \rowcolor{cyan!30}
    AT-EDM-d & 27.23 & 0.3245 & 4.5 \\
    \bottomrule
  \end{tabular}
  \vspace{-1em}
\end{table}

\begin{figure}[t]
  \centering
  % \vspace{-1em}
   \includegraphics[width=\linewidth]{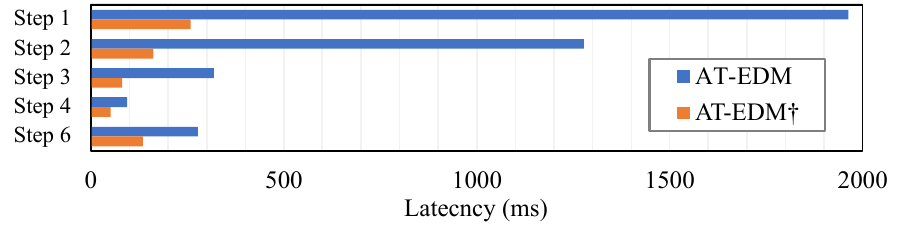}
   \vspace{-2.25em}
   \caption{Latency incurred by different pruning steps shown in Fig. \ref{fig:overview}. Measured w/ FO under CI. Note that under DI, the latency of Step 1 (get the attention map) is eliminated.
   }
   % \vspace{-1.5em}
   \label{fig:prune_bd}
\end{figure}

\noindent\textbf{Latency Analysis.} SD-XL uses the Fused Operation (FO) library, xformers \cite{xFormers}, to boost its generation. The Current Implementation (CI) of xformers does not provide attention maps as intermediate results; hence, we need to additionally calculate the attention maps. We discuss the sampling latency for three cases: (I) without FO, (II) with FO under CI, and (III) with FO under the Desired Implementation (DI), which provides attention maps as intermediate results. Table \ref{tab:latency} shows that with FO, the cost of deploying pruning at the second feature level exceeds the latency reduction it leads to. Hence, AT-EDM$^\dagger$ is faster than AT-EDM. Fig.~\ref{fig:prune_bd} shows the extra latency incurred by different pruning steps shown in Fig. \ref{fig:overview}. With a negligible quality loss, \textbf{AT-EDM achieves 52.7\%, 15.4\%, 17.6\% speed-up in terms of latency w/o FO, w/ FO under CI, w/ FO under DI, respectively}, which outperforms the state-of-the-art work by a clear margin. We present the memory footprint of AT-EDM in Supplementary Material.

% For AT-EDM (AT-EDM$^\dagger$) w/ FO, the latency of steps shown in Fig. 3 of the main paper: [Step 1] 2.0s (0.26s); [Step 2] 1.3s (0.16s); [Step 3] 0.3s (0.081s); [Step 4] 0.09s (0.051s); [Step 6] 0.3s (0.13s). 

% \vspace{-1em}

\begin{table}[t]
  \small
  % \footnotesize
  % \scriptsize
  \centering  
  \caption{Comparison between sampling latency in different cases. $^\dagger$ means not deploying pruning at the second feature level.}
  \vspace{-1em}
    \begin{tabular}{ccccc}
    \toprule
    %\multicolumn{2}{c}{Part}                   \\
    %\cmidrule(r){1-2}
    Model  &  SD-XL    & ToMe & AT-EDM & AT-EDM$^\dagger$ \\
    Ave. FLOPs/step&  6.7 T & 4.1 T & 4.1 T & 4.5 T \\
    \midrule
    w/o FO & 31.0s & 21.0s & \textbf{20.3s} & 22.1s  \\
    w/ FO under CI   & 18.0s  & 17.7s  & 18.3s & \textbf{15.6s} \\
    w/ FO under DI & 18.0s  & 17.7s & 16.3s & \textbf{15.3s} \\
    \bottomrule
    \end{tabular}%
  \label{tab:latency}%
  \vspace{-0.75em}
\end{table}%

\subsection{Ablation Study}
\label{sec:ablation}

\begin{figure}[t]
  \centering
  % \vspace{-1em}
   \includegraphics[width=\linewidth]{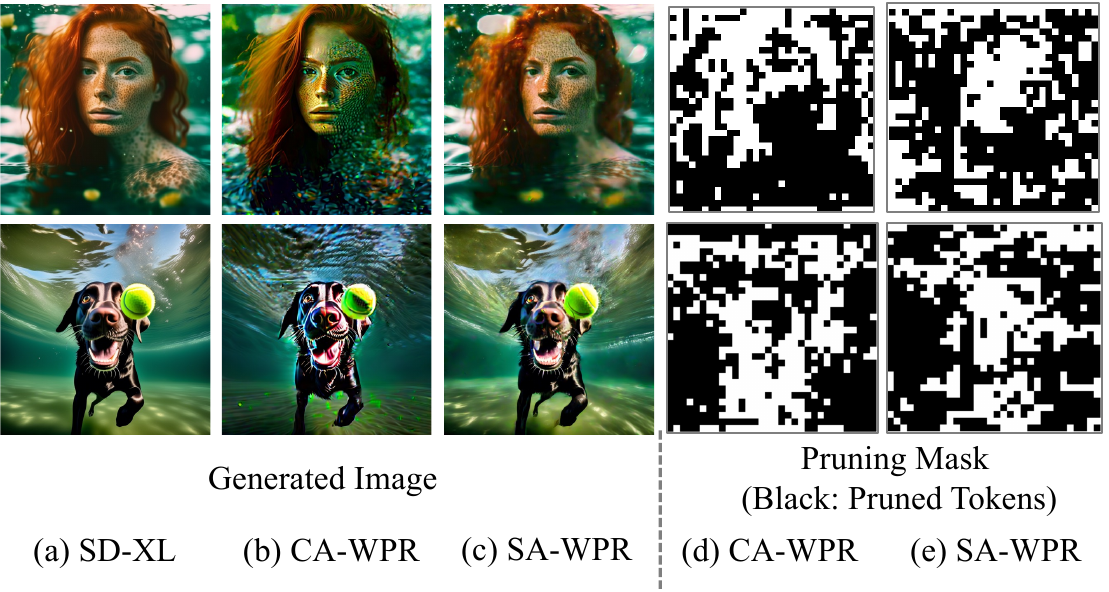}
   \vspace{-2em}
   \caption{Comparison between different implementations of G-WPR: CA-based WPR and SA-based WPR. 
   In general, CA-based WPR may remove too many background tokens, making the background not recoverable, 
   while SA-based WPR preserves the image quality better.
   }
   \vspace{-1em}
   \label{fig:comp_casa}
\end{figure}

\noindent\textbf{Self-Attention (SA) vs. Cross-Attention (CA).} G-WPR can potentially use attention maps from self-attention (SA-based WPR) and cross-attention (CA-based WPR). We provide a detailed comparison between the two implementations. We visualize their pruning masks and provide generated image examples for a visual comparison in Fig.~\ref{fig:comp_casa}. This figure indicates that SA-based WPR outperforms CA-based WPR. The reason is that CA-based WPR prunes too many background tokens, making it hard to recover the background via similarity-based copy.

\begin{figure}[t]
  \centering
  % \vspace{-1em}
   \includegraphics[width=\linewidth]{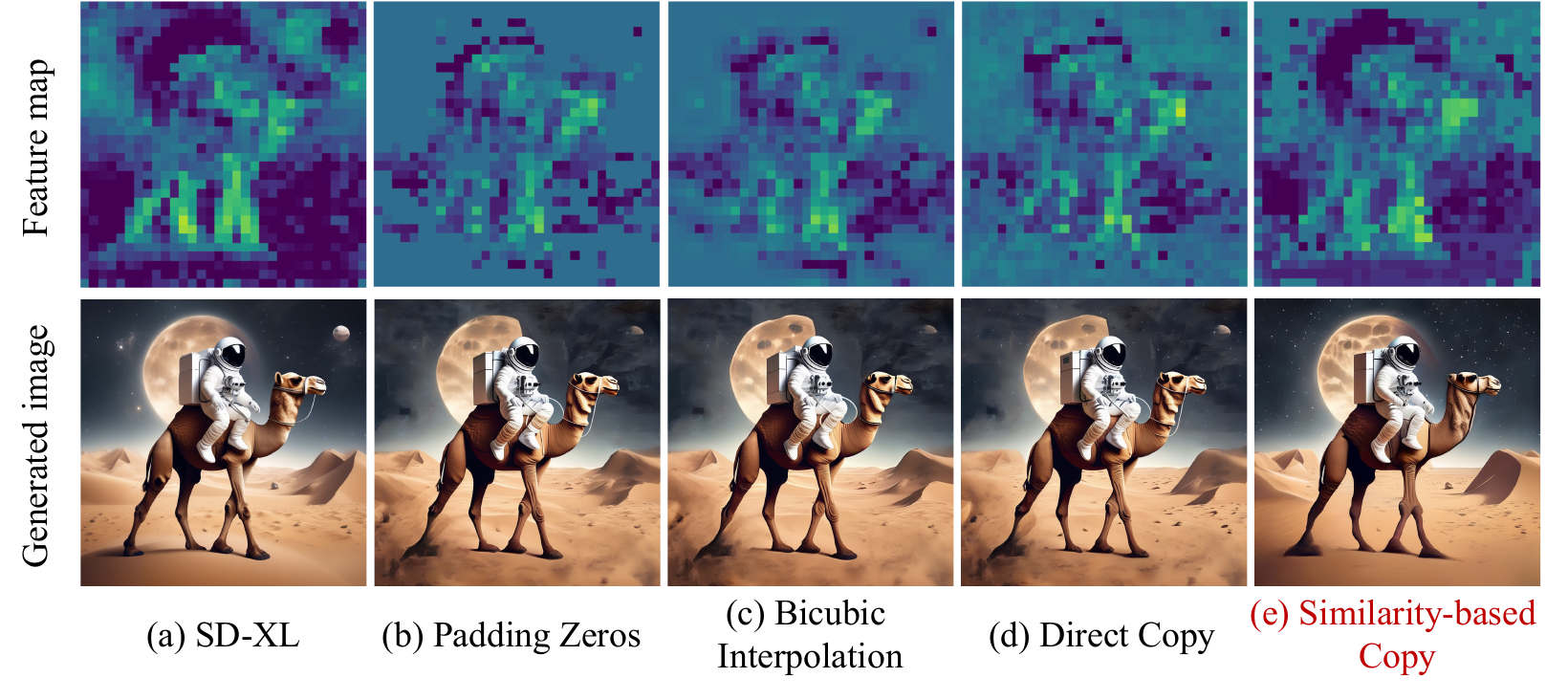}
   \vspace{-2em}
   \caption{Different methods to recover the pruned tokens. 
   Zero padding (\textbf{Col.~b}), bicubic interpolation (\textbf{Col.~c}), and direct copy (\textbf{Col.~d}) can hardly recover pruned tokens and result in noticeable image degradation with blurry background (incomplete moon).
   On the other hand, similarity-based copy (\textbf{Col.~e}) provides better image quality and keeps the complete moon in the original image. Better viewed when zoomed in.
   }
   \vspace{-1em}
   \label{fig:comp_fill}
\end{figure}

\noindent\textbf{Similarity-based Copy.} We provide comparisons between different methods to fill the pruned pixels in Fig.~\ref{fig:comp_fill}, which demonstrate the advantages of our similarity-based copy method. Images generated by bicubic interpolation are quite similar to those generated by padding zeros because interpolation usually assigns near-zero values to pruned tokens that are surrounded by other pruned tokens and can hardly recover them. %(Show feature maps here or in Supplementary Materials) 
Direct copy means directly copying corresponding token values before the first pruning layer in the attention block to recover the pruned tokens, where the following attention layers do not process the copied values. Thus, the copied values cannot recover the information in pruned tokens and even negatively affect the retained tokens. On the contrary, similarity-based copy uses attention maps and tokens that are retained to recover the pruned tokens, providing significantly higher image quality.

\begin{figure}[t]
  \centering
   \includegraphics[width=\linewidth]{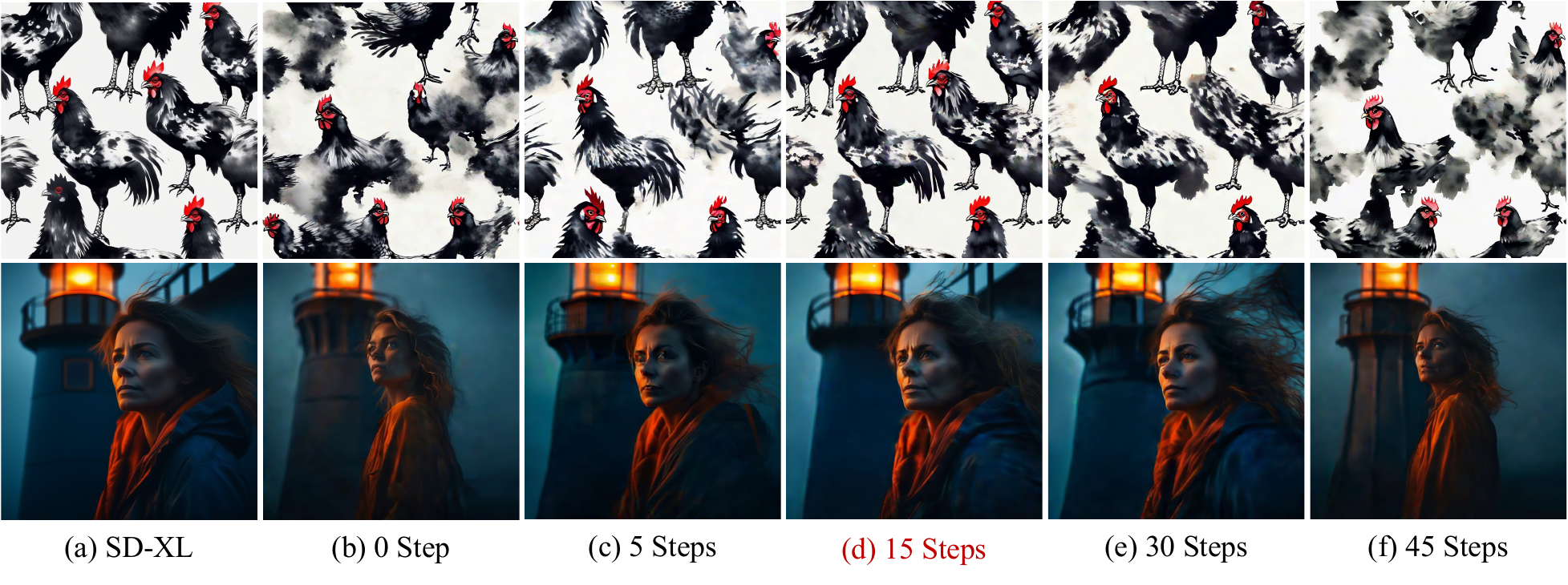}
   \vspace{-2em}
   \caption{Comparison between different numbers of early prune-less steps where 0 step is the same as without DSAP. 
   We find that pruning less on the first 15 steps achieves the best quality.}
   \vspace{-1em}
   \label{fig:comp_earlynum}
\end{figure}

\noindent\textbf{Denoising-Steps-Aware Pruning.} We explore different design choices for DSAP. 

\noindent(1) The prune-less schedule selects one attention block from each down-stage and up-stage in the U-Net and skips the token pruning in it. According to ablation results shown in Supplementary Material, F-L (First-Last) appears to be the best one, i.e., leaving the first attention block of down-stages and the last attention block of up-stages unpruned in early denoising steps. 

\noindent(2) We then explore how the number of early prune-less denoising steps
affects the generated image quality in Fig.~\ref{fig:comp_earlynum}. Note that we keep the FLOPs budget fixed and adjust the pruning rate accordingly when we change the number of prune-less steps. 
% We fix the total number of sampling steps to 50. 
This figure shows that the setting of 15 early prune-less steps provides the best image quality. 
% We further validate this choice by analyzing the variance of attention maps from different denoising steps and their ability to identify unimportant tokens in Supplementary Material. 
Note that the setting of zero prune-less step is identical to the setting without DSAP, and 5, 15, 30, 45 prune-less steps represents setting the boundary in Region I, II, III, IV of Fig.~\ref{fig:var_attn}, respectively. The results indicate that placing the boundary between the prune-less and normal schedule in Region II performs best. This meets our expectation because the variance of attention maps becomes high enough to identify unimportant tokens well in Region II.

% However, this method meets two severe problems when it is applied to state-of-the-art model, stable diffusion XL. First, the quality of generated images is very bad. In terms of quantitative metrics, as shown in Fig. \ref{fig:tome_fid}, FID scores increase a lot and CLIP scores decrease a lot after ToMe is applied. This means the generated images have not only lower fidelity but also worse text-image alignment. 

\section{Conclusion}
\label{sec:conclusion}

In this article, we proposed AT-EDM, a novel framework for accelerating DMs at run-time without retraining.
AT-EDM has two components: a single-denoising-step token pruning algorithm and a cross-step pruning schedule (DSAP). 
In the single-denoising-step token pruning, AT-EDM 
exploits attention maps in pre-trained DMs to identify unimportant tokens and prunes them to accelerate the generation process.
To make the pruned feature maps compatible with the latter convolutional blocks, AT-EDM again uses attention maps to reveal similarities between tokens and copies similar tokens to recover the pruned ones.
DSAP further improves the generation quality of AT-EDM. We find such a pruning schedule can also be applied to other methods like ToMe.
Experimental results demonstrate the superiority of AT-EDM with respect to image quality and text-image alignment compared to state-of-the-art methods. 
Specifically, on SD-XL, AT-EDM achieves a 38.8\% FLOPs saving and up to 1.53$\times$ speed-up while obtaining nearly the same FID and CLIP scores as the full-size model, outperforming prior art. 

\section*{Acknowledgment} 
This work was supported in part by an Adobe summer internship and in part by NSF under Grant No. CCF-2203399.

% \input{sec/6_acknowledgment}

% \newpage
% \null
\newpage

{
    \small
    \bibliographystyle{ieeenat_fullname}
    \bibliography{main}
}

\clearpage
\maketitlesupplementary
\appendix

The Supplementary Material is organized as follows. We first provide more implementation details of AT-EDM in Section \ref{app:implementation}, including a detailed illustration of the SD-XL backbone. Then, we provide a more comprehensive comparison with the state-of-the-art method, ToMe~\cite{tomesd}, in Section \ref{app:comptome}, including an analysis of why ToMe performs worse on SD-XL~\cite{sdxl} than on previous versions of Stable Diffusion Models (SDMs). We provide more ablation results in Section \ref{app:ablation} to justify our design choices in the main article. We analyze the memory footprint of AT-EDM in Section \ref{app:memory}. AT-EDM is orthogonal to various efficient DM methods, such as sampling distillation, thus can further boost their efficiency. To support this claim, we deploy AT-EDM in the distilled version of SD-XL, SDXL-Turbo\footnote{https://huggingface.co/stabilityai/sd-turbo}, and show corresponding experimental results in Section \ref{app:stack}. We discuss limitations and trade-offs of AT-EDM in Section \ref{app:limitation} and potential negative social impacts of AT-EDM in Section \ref{app:negimpact}.

\section{Implementation Details}

\label{app:implementation}

In this section, we provide more details of the implementation of AT-EDM. We first introduce the architecture of our SD-XL backbone as background material and then describe our single-step and cross-step pruning schedules in detail. We describe details of the evaluation and our calibration block for FLOPs measurement in the end.

\subsection{The SD-XL Backbone}

The state-of-the-art version of SDM is SD-XL. Compared with previous versions of SDM, it increases the quality of generated images significantly. Thus, we select SD-XL as the backbone model in this article. Specifically, we deploy AT-EDM and ToMe on \texttt{SDXL-base-0.9}. 
% We show the U-Net architecture of SD-XL in Fig.~\ref{fig:sdxl_unet}. 
The architecture has two main differences from that of previous SDMs, such as SD-v1.5 and SD-v2.1: (1) attention blocks at the highest feature level (i.e., with the most tokens) are deleted; (2) attention blocks can potentially include multiple attention layers (an attention layer is composed of self-attention, cross-attention, and feed-forward network), such as A2 (includes 2 attention layers) and A10 (includes 10 attention layers).

\begin{figure}[h]
  \centering
   \includegraphics[width=\linewidth]{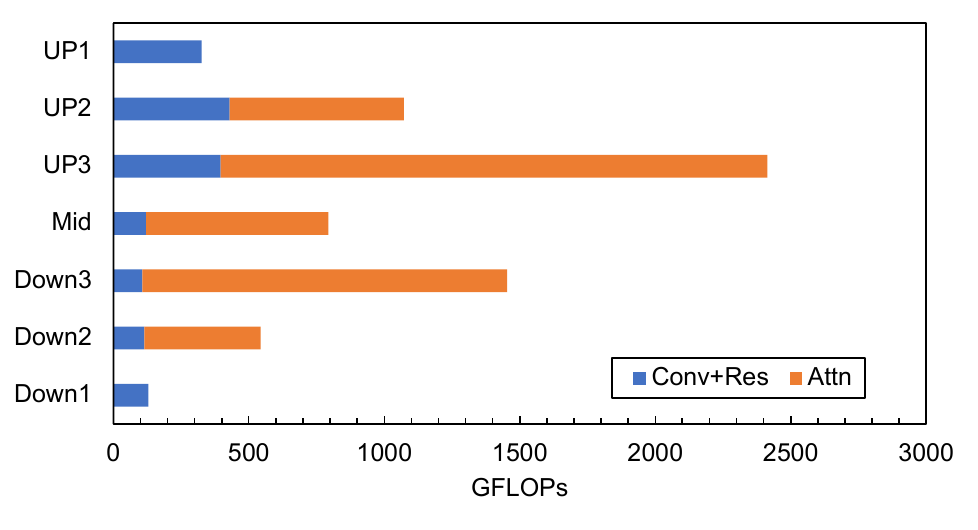}

   \caption{The FLOPs breakdown of SD-XL. Measured with 1024$\times$1024 px image generation.}
   \label{fig:flops_xl}
\end{figure}

\begin{figure}[h]
  \centering
   \includegraphics[width=\linewidth]{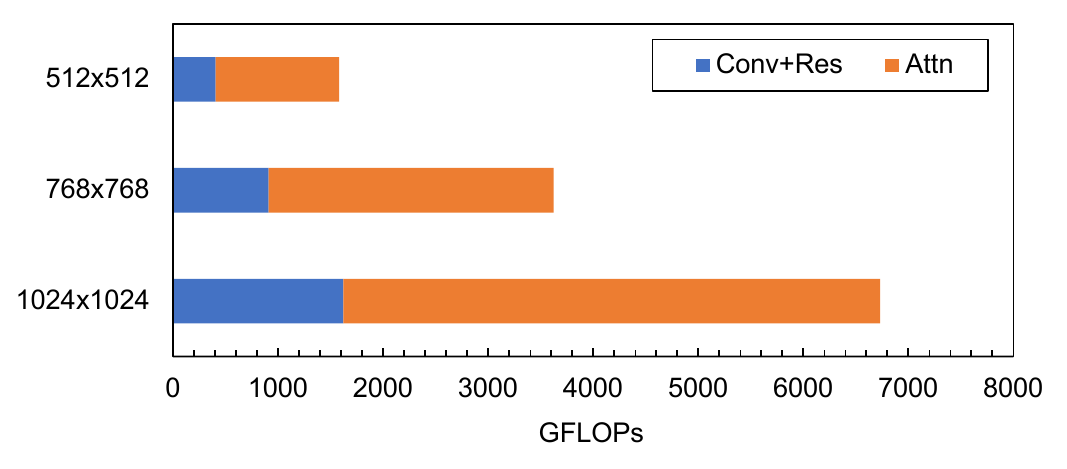}
   \caption{The FLOPs breakdown of ResNet blocks and attention blocks in SD-XL at different image resolutions.}
   \label{fig:sdxl_scale_flops}
\end{figure}

To validate the conclusion that the cost of attention layers dominates the sampling cost, we investigate the FLOPs cost of SD-XL. Its FLOPs profile is shown in Fig.~\ref{fig:flops_xl}. This figure indicates that the attention block dominates the computational cost of all stages that include attention. We also investigate the scaling law of SD-XL at different generation resolutions, as shown in Fig. \ref{fig:sdxl_scale_flops}. We observe that the attention block dominates the cost at all resolutions. Note that the FLOPs cost of attention blocks does not scale much faster than that of ResNet blocks when the generation resolution increases. We believe this is due to the elimination of attention blocks at the highest feature level and the addition of attention layers at the lowest feature level, making the cost of feed-forward layers, which scales linearly with an increment in token numbers, a huge part of the cost of attention layers.

\subsection{Pruning in a Single Denoising Step}

For a concise design, we always insert the pruning layer after the first attention layer of each attention block. All the other attention layers in this attention block can benefit from the reduction in token numbers. We may also insert multiple pruning layers at various locations in an attention block, which prunes tokens gradually. However, this requires a more thorough hyperparameter search to ensure a good balance between FLOPs cost and image quality.

\subsection{The Prune-Less Schedule}
\label{app:lessprune}

\begin{figure*}[h]
  \centering
   \includegraphics[width=\linewidth]{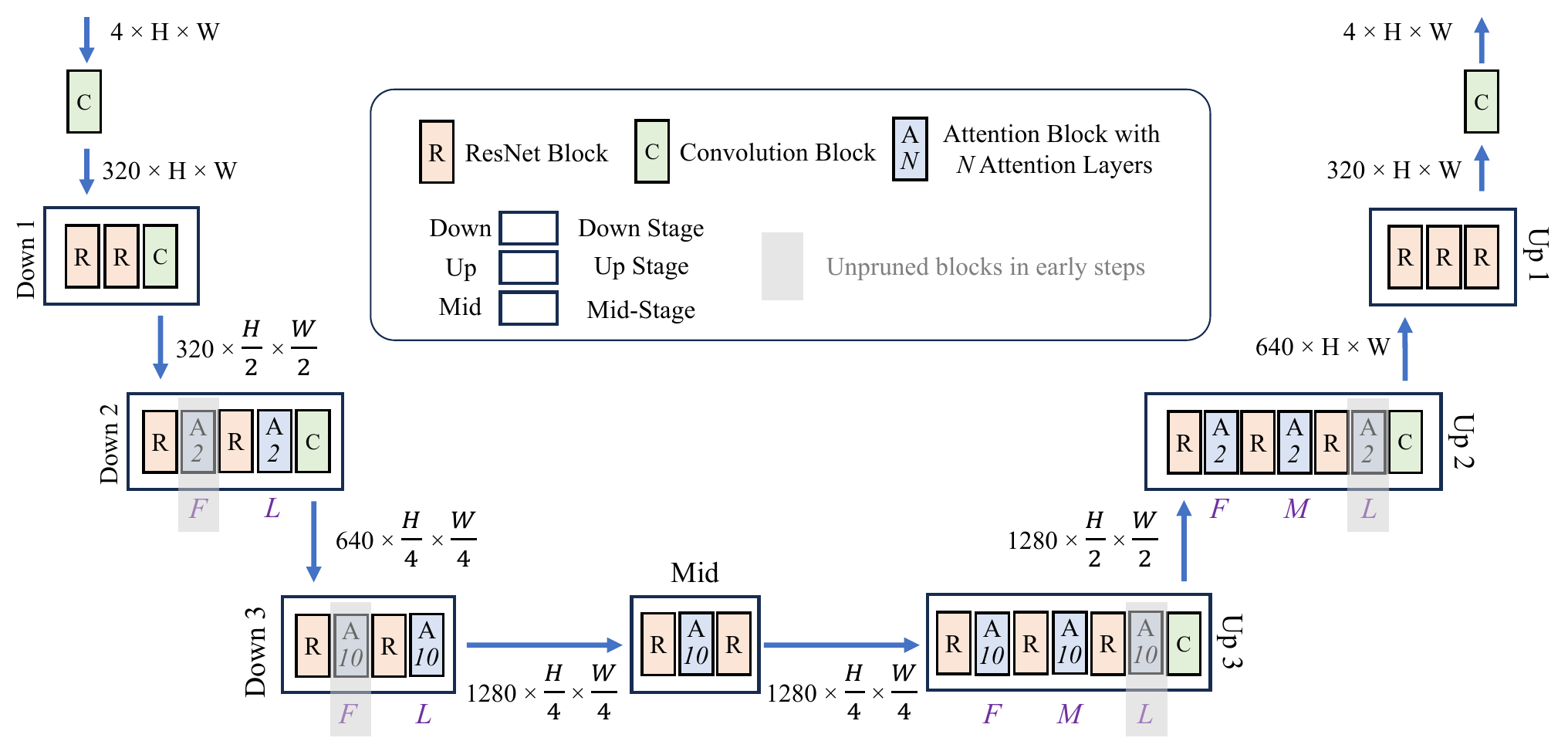}
   \caption{The U-Net architecture of SD-XL. Residual connections are not shown here for brevity. The example in this figure generates a $8H\times 8W$ pixel image. The input/output size of each stage is shown in the $C\times H\times W$ format, where $C$ is the number of channels; $H$ and $W$ represent the resolution. There are two attention blocks \{F(First), L(Last)\} in each downsampling stage and three \{F(First), M(Middle), L(Last)\} in each upsampling stage. In the prune-less schedule, we do not apply pruning to attention blocks in the gray rectangles. Downsampling stage 1, 2, and 3 is at the first, second, and third feature level, respectively. AT-EDM$^\dagger$ does not apply pruning to attention blocks at the second feature level.}
   \label{fig:sdxl_unet}
\end{figure*}

Early denoising steps determine the layout of the generated images and have a weaker ability to differentiate between unimportant tokens \cite{prompt2prompt}. Thus, we need heterogeneous denoising steps and, hence, use a less aggressive pruning schedule for some of the early denoising steps. 
% The less-aggresive pruning schedule is shown in Fig.~\ref{fig:sdxl_unet}.

In the normal pruning setting, when we target 4.1 TFLOPs for each sampling step, we use a pruning rate of 63\% (i.e., retain 37\% tokens) after the first attention layer of A2 and A10; in the prune-less schedule, we do not apply pruning to attention blocks in the gray rectangles shown in Fig. \ref{fig:sdxl_unet}. We validate the choice of not deploying pruning through ablative experimental results shown in the main article.

\subsection{Details of Evaluation}

When measuring the FID and CLIP scores on MS-COCO 2017 \cite{mscoco}, we deduplicate captions to make sure each image corresponds to a single caption. We center cropped images in the validation set, resize them to 1024$\times$1024 px, and use the \texttt{clean-fid} library\footnote{https://github.com/GaParmar/clean-fid/tree/main} to calculate FID scores. We use the ViT-G/14 model of Open-CLIP\footnote{https://github.com/mlfoundations/open\_clip} to calculate the CLIP scores of generated images. We set the batch size to 3 when we generate images for visual comparison and quantitative analysis. We run all experiments on a single NVIDIA A100-40GB GPU.

\subsection{Calibration Block for FLOPs Measurement}

The popular library for FLOPs measurement, \texttt{fvcore}\footnote{https://github.com/facebookresearch/fvcore}, is not natively compatible with SDMs. Thus, we use the \texttt{THOP}\footnote{https://github.com/Lyken17/pytorch-OpCounter} library instead to measure the FLOPs cost of SDMs. However, we found it does not correctly compute the FLOPs cost of self-attention. The FLOPs cost of sampling steps given by this library scales linearly as the number of image tokens. This is unreasonable because the cost of self-attention in sampling steps scales quadratically when the number of tokens increases (other parts of a sampling step scale linearly). After a thorough investigation of the behavior of \texttt{THOP}, we found it basically does not take the cost of self-attention into account. Thus, we design a calibration block to supplement the missed term of FLOPs cost for each attention block:

\begin{equation}
    F_{cali} = 4 \times B \times N_a \times (HW)^2 \times C
\end{equation}

\noindent
where $B$ is the batch size; $N_a$ is the number of attention layers in this attention block; $HW$ is the number of image tokens; and $C$ is the number of channels. The factor $4$ is due to the fact that (1) there are two images processed at the same time for each generated image in a batch (one is guided by the prompt, and another is not); (2) there are two Matrix-Matrix Multiplications (MMMs) in self-attention.
\section{Comprehensive Comparison with ToMe}

\label{app:comptome}

In this section, we first analyze why ToMe cannot replicate on SD-XL its good performance on previous SDMs in Section \ref{app:profile}. Then, 
we provide complete FID-CLIP curves to compare AT-EDM with ToMe in Section \ref{app:complete-curve}. In the end, 
we present cases in which both AT-EDM and ToMe perform well and visually compare AT-EDM and ToMe under various FLOPs budgets in Section \ref{app:examples}. 

\subsection{Deploying ToMe on SD-XL}
\label{app:profile}

\begin{figure}[t]
  \centering
   \includegraphics[width=\linewidth]{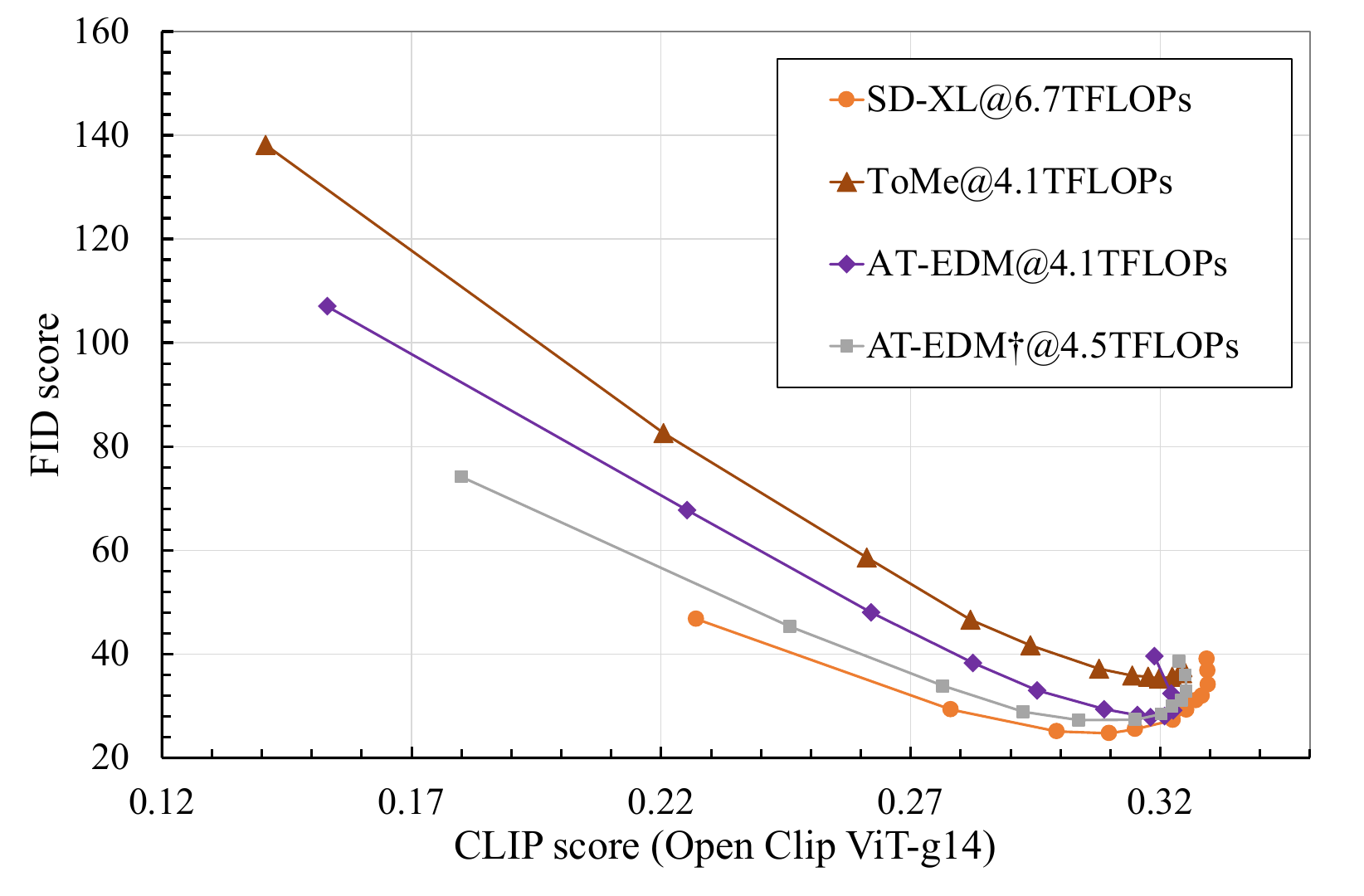}
   \vspace{-2.5em}
   \caption{Complete FID-CLIP score curves. The used CFG scales are [1.0, 1.5, 2.0, 2.5, 3.0, 4.0, 5.0, 6.0, 7.0, 9.0, 12.0, 15.0].}
   \vspace{-1em}
   \label{fig:tome_fid_complete}
\end{figure}

For SD-v1.x and SD-v2.x, ToMe maintains the generated image quality quite well after token merging. However, as we demonstrate in the main article, ToMe incurs obvious quality degradation on SD-XL after token merging.

In the default setting of ToMe, it only merges tokens for attention blocks at the highest feature level and their self-attention. However, SD-XL eliminates attention blocks at the highest abstraction level and native ToMe does not do anything to this backbone. Thus, it is necessary to expand its merging range to \textbf{attention blocks at all feature levels}. In addition, since SD-XL adds a lot more attention layers at the lowest feature level, where tokens are significantly fewer than at higher feature levels, self-attention no longer dominates the cost of attention layers. Given that the merging ratio of ToMe has an upper limit of 75\%, it is not enough to only merge tokens for self-attention to meet the 4.1 TFLOPs budget. Thus, it is necessary to expand its merging range to \textbf{Cross-Attention (CA), Self-Attention (SA), and the Feed-Forward (FF) network}. We believe the expanded deployment range of token merging leads to the relatively poor performance of ToMe on SD-XL. Note that to meet the 4.1 TFLOPs budget for each sampling step, we set the merging ratio to 50\% for ToMe under the expanded merging range.

\subsection{Complete FID-CLIP Curves}
\label{app:complete-curve}

We explore the trade-off between the CLIP and FID scores through various CFG scales. We show the complete FID-CLIP curves in Fig.~\ref{fig:tome_fid_complete}. AT-EDM$^\dagger$ does not deploy pruning at the second feature level (as mentioned in the caption of Fig.~\ref{fig:sdxl_unet}). This figure illustrates that for most CFG scales, AT-EDM not only lowers the FID score but also results in higher CLIP scores than ToMe, implying that images generated by AT-EDM not only have better quality but also better text-image alignment.

\begin{figure}[t]
  \centering
   \includegraphics[width=\linewidth]{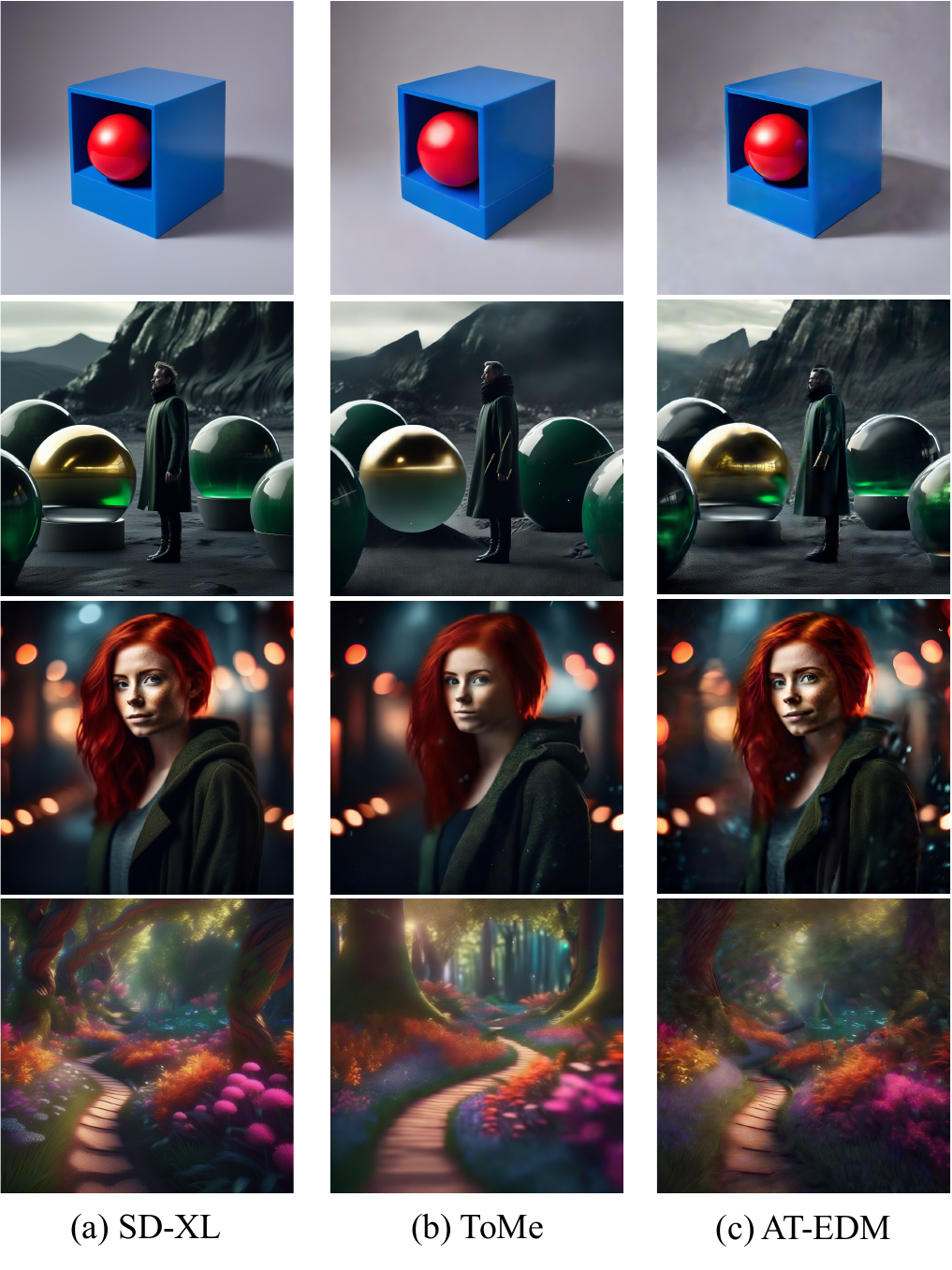}
   \caption{Examples on which both AT-EDM and ToMe perform well. Each row of this figure corresponds to the following typical cases: (1) simple single main object with a simple background; (2) multiple main objects; (3) complex single main object; (4) complex scene without a main object.}
   \label{fig:comp_good}
\end{figure}

\subsection{More Images from AT-EDM and ToMe}
\label{app:examples}

\begin{figure*}[h]
  \centering
   \includegraphics[width=\linewidth]{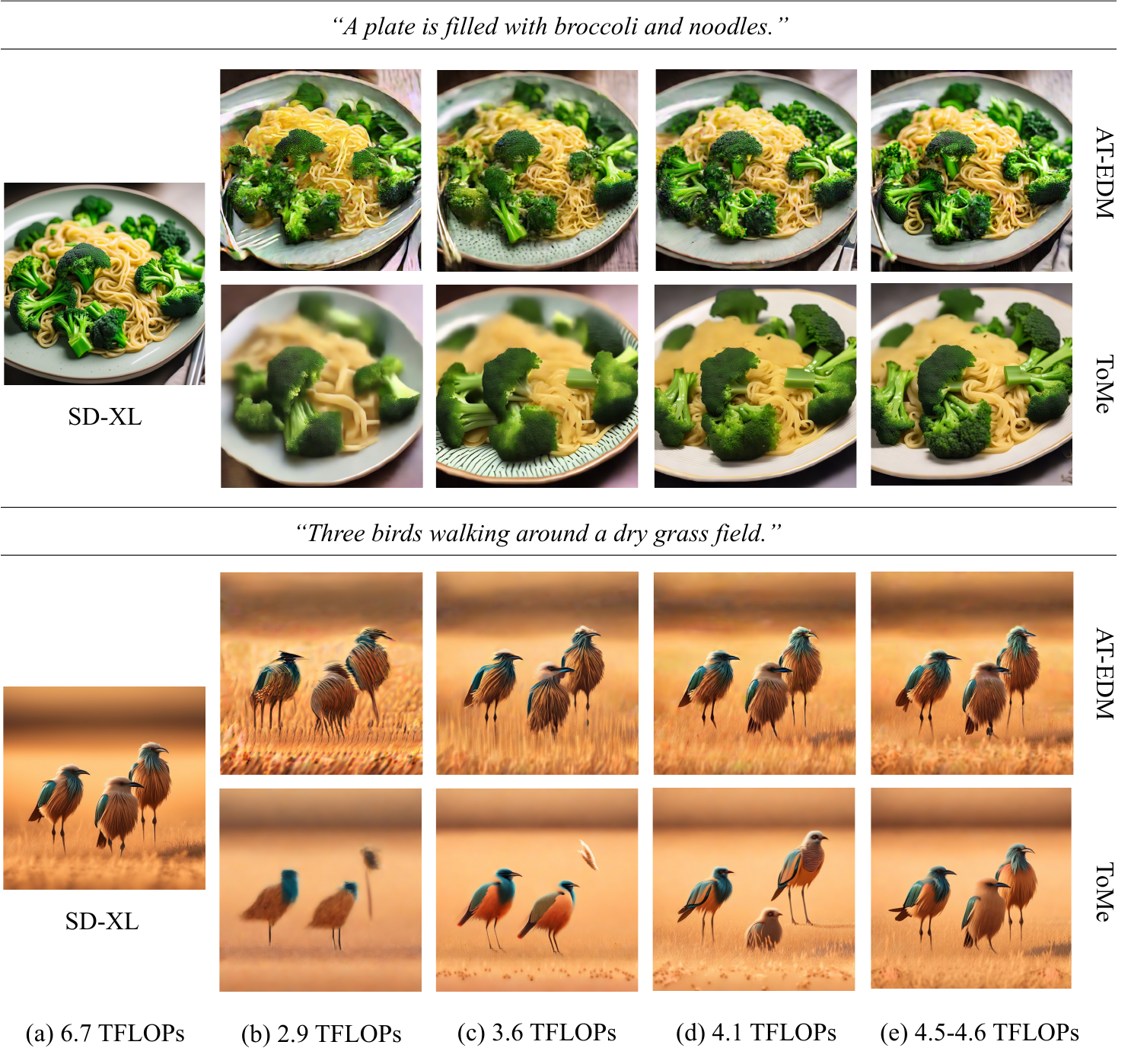}
   \vspace{-2.5em}
   \caption{Comparison between AT-EDM and ToMe under different FLOPs budgets. Note that for \textbf{Col.e}, the average cost of each sampling step for AT-EDM (ToMe) is 4.52 (4.56) TFLOPs. Prompts are selected from the MS-COCO 2017 validation dataset.}
   % \vspace{-1em}
   \label{fig:comp_flops}
\end{figure*}

In some cases, ToMe performs fairly well and has its merits. We present several typical examples in Fig.~\ref{fig:comp_good}. The first example in the first row represents the case of a simple main object with a simple background. Both ToMe and AT-EDM preserve the main object quite well. The second row represents a more complex case in which there are multiple main objects in the generated image. Although ToMe loses some texture details, it preserves the overall layout quite well. The third row is the case of a typical complex main object, a human face. In this example, ToMe preserves the face without artifacts. The last row of this figure demonstrates the case of generating a complex scene without a main object. In this case, both ToMe and AT-EDM can maintain the layout well while supplementing some details. These examples show that ToMe is a strong baseline and it is non-trivial to outperform it.

We also provide visual examples of ToMe and AT-EDM under different FLOPs budgets in Fig.~\ref{fig:comp_flops}. It indicates that AT-EDM outperforms ToMe under any FLOPs budget. We also observe that AT-EDM needs at least 3.6 TFLOPs budget to ensure an acceptable image quality.
\section{More Ablation Experiments}

In this section, we supplement ablation experiments to validate our design choices. We first discuss the deployment location for run-time pruning and then compare different implementations of the mapping function $f(\mathbf{A},s_K)$ for CA-based WPR. Note that CA-based WPR and SA-based WPR are two implementations of G-WPR and we mainly focus on CA-based WPR in this section. We also investigate the schedule that prunes more in early denoising steps and verify our intuition of pruning less in early steps.

\label{app:ablation}

\subsection{Deployment Location for Run-Time Pruning}

\begin{figure}[t]
  \centering
   \includegraphics[width=\linewidth]{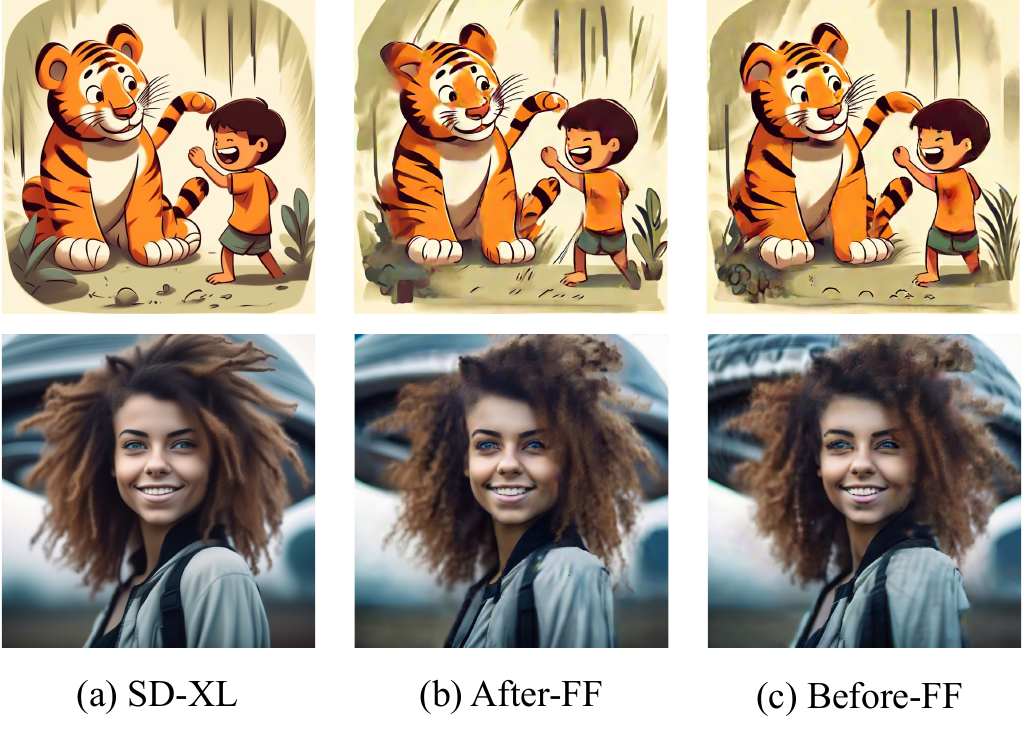}

   \caption{Comparison between inserting the pruning layer after the FF and before the FF layer.}
   \label{fig:comp_ff}
\end{figure}

In our default setting, we use generated masks after the FF layer to perform token pruning. Another option is to perform pruning early before the FF layers, which results in a little bit of extra FLOP savings at the cost of image quality. We provide several visual examples in Fig.~\ref{fig:comp_ff}. Note that here, we simply change the pruning layer insertion location without keeping the total FLOPs cost fixed, which is different from what we do in the ablation experiments in the main article. We find that inserting the pruning layer before the FF layer indeed hurts image quality (although slightly). For example, the plant in the first example and the UFO in the second example become worse. Given that pruning before the FF layer only results in marginal extra FLOPs savings (reduces the cost from 4.1 TFLOPs to 4.0 TFLOPs), we choose to prune after the FF layer to obtain better image quality.

\subsection{Implementations of CA-based WPR}

To generalize WPR to cross-attention, we need to design a function $f(\mathbf{A},s_K)$ that maps the importance of Key tokens to that of Query tokens. The intuition behind designing this function is that vital Query tokens should devote much of their attention to important Key tokens. Thus, the desired attention distribution should satisfy: (1) similarity to the importance distribution of Key tokens; (2) concentration on a few tokens. Then, when designing $f(\mathbf{A},s_K)$, we need to (1) reward the similarity between the attention distribution (i.e., each row of $\mathbf{A}$) and the importance distribution (i.e., $s_K$); (2) penalize uniform attention distribution. Based on these points, we obtain several implementations of $f(\mathbf{A},s_K)$. We had mentioned an entropy-based implementation in the main article, which rewards similarity through the dot-product and penalizes uniform distribution through entropy. We provide additional implementations here: 

\noindent
(I) \textbf{Hard-clip}-based implementation

\begin{equation}
    s^{t+1}_Q(x_i) = f(\mathbf{A}, s^{t+1}_K)=\sum_{j=1}^{N}\epsilon(A_{i,j}-\eta)\cdot s^{t+1}_K(x_j)
\end{equation}

\noindent
where $\epsilon(x)=1$ if $x\geq0$, $\epsilon(x)=0$ if $x<0$; $\eta$ is the threshold of attention (we set it to 0.2 as the default setting); $A_{i,j}$ is the attention from Query $q_i$ to Key $k_j$. 

\noindent
(II) \textbf{Soft-clip}-based implementation

\begin{equation}
    s^{t+1}_Q(x_i) = f(\mathbf{A}, s^{t+1}_K)=\sum_{j=1}^{N}\text{Sig}(A_{i,j}-\eta)\cdot s^{t+1}_K(x_j)
\end{equation}

\noindent
where $\text{Sig}(x)=\frac{1}{1+e^{-x}}$. 

\vspace{1em}

\noindent
(III) \textbf{Power}-based implementation

\begin{equation}
    s^{t+1}_Q(x_i) = f(\mathbf{A}, s^{t+1}_K)=\sum_{j=1}^{N}(\beta\cdot s^{t+1}_K(x_j))^ {\alpha\cdot A_{i,j}}
\end{equation}

\noindent
where $\alpha$ and $\beta$ are scaling factors to ensure that $\beta\cdot s^{t+1}_K(x_j)>1$ and $\alpha\cdot A_{i,j}>1$ for large $s^{t+1}_K(x_j)$ and $A_{i,j}$. Here, we let $\alpha=5$ and $\beta=\frac{N_t}{2}$, where $N_t$ denotes the number of Key tokens.

\begin{figure*}[h]
  \centering
   \includegraphics[width=\linewidth]{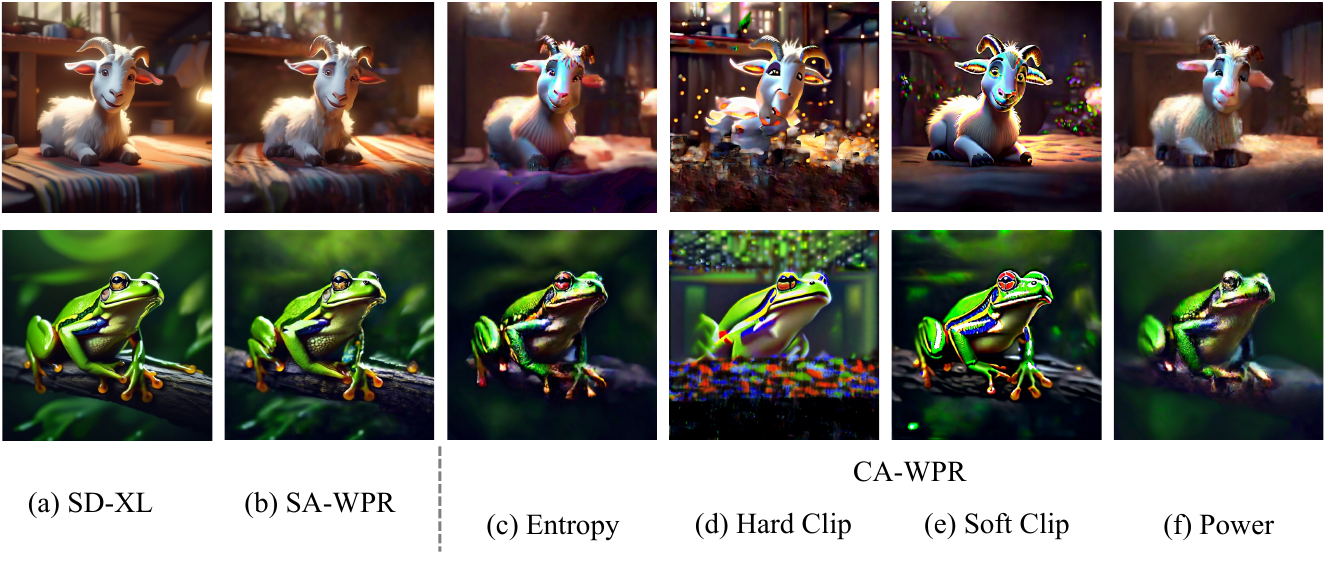}

   \caption{Comparison between different implementations of Cross-Attention-based WPR. None of them can outperform Self-Attention-based WPR.}
   \label{fig:comp_caimp}
\end{figure*}

We compare these implementations visually in Fig.~\ref{fig:comp_caimp}. We find that among these implementations, the hard-clip-based implementation performs the worst. Although the entropy-based implementation and the power-based implementation are better than other implementations for CA-based WPR, none of them can outperform SA-based WPR. Thus, we use SA-based WPR as our default setting in AT-EDM.

\subsection{Prune-Less Schedule for Early Denoising Steps}

\begin{figure*}[t]
  \centering
  % \vspace{-2em}
   \includegraphics[width=\linewidth]{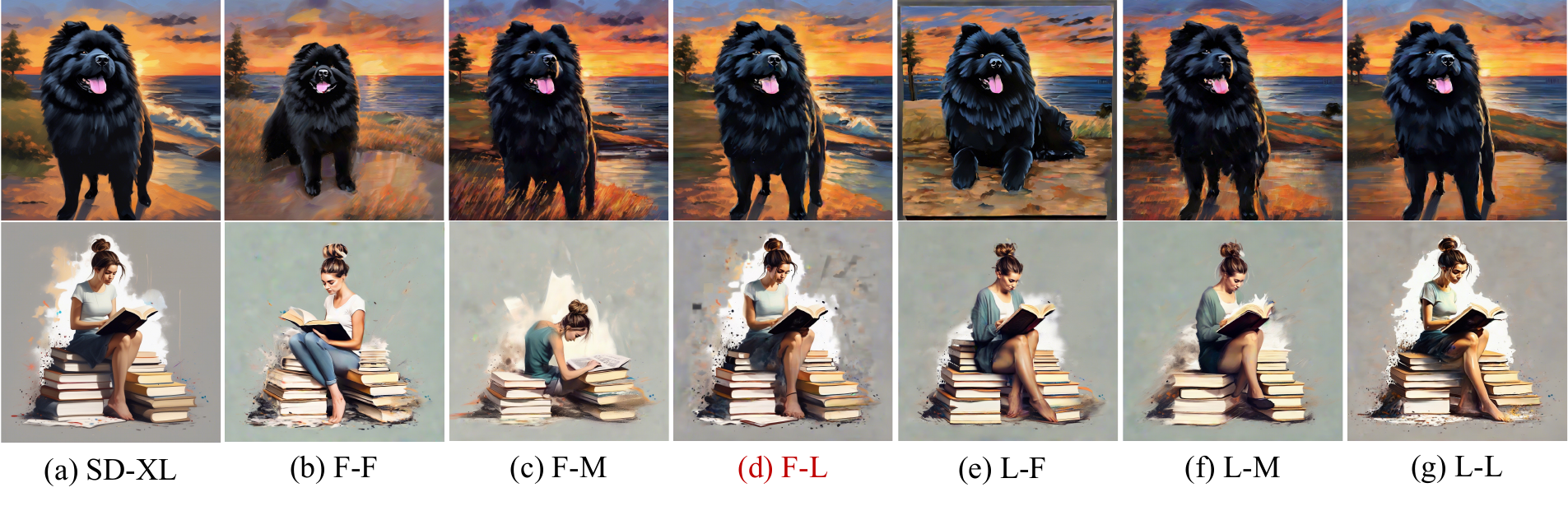}
   % \vspace{-2.5em}
   \caption{Comparison between different prune-less settings. There are two attention blocks \{F(First), L(Last)\} that are left unpruned in the downsampling stages and three \{F(First), M(Middle), L(Last)\} in the upsampling stages. 
   Results indicate that F-L is the best schedule.}
   % \vspace{-1.5em}
   \label{fig:comp_retain}
\end{figure*}

The prune-less schedule selects one attention block from each down-stage and up-stage in the U-Net and skips the token pruning in it. We generate images with the same prompts and different selections, as shown in Fig.~\ref{fig:comp_retain}. It indicates that F-L appears to be the best choice. F-L is the schedule that we show in Fig.~\ref{fig:sdxl_unet}.

\subsection{The Number of Prune-Less Steps}

The intuitions that we use to design the prune-less schedule in the early denoising steps are (1) early denoising steps determine the layout of generated images and thus are crucial; (2) early denoising steps have a weaker ability to differentiate unimportant tokens. The first intuition prohibits us from pruning more tokens in the early steps (see Section \ref{app:prune_more}). The second intuition guides us to choose the number of prune-less steps. The variance of attention maps reflects their ability to differentiate unimportant tokens since the attention score of unimportant tokens deviates significantly from that of normal tokens. We show the variance of attention maps given by different denoising steps in Fig.~\ref{fig:var_attn}. The figure indicates that the variance is more than 1.0E-5 after the first 15 denoising steps. This supports our hyperparameter choice.

\subsection{Prune More in Early Denoising Steps}
\label{app:prune_more}

\begin{figure}[t]
  \centering
   \includegraphics[width=\linewidth]{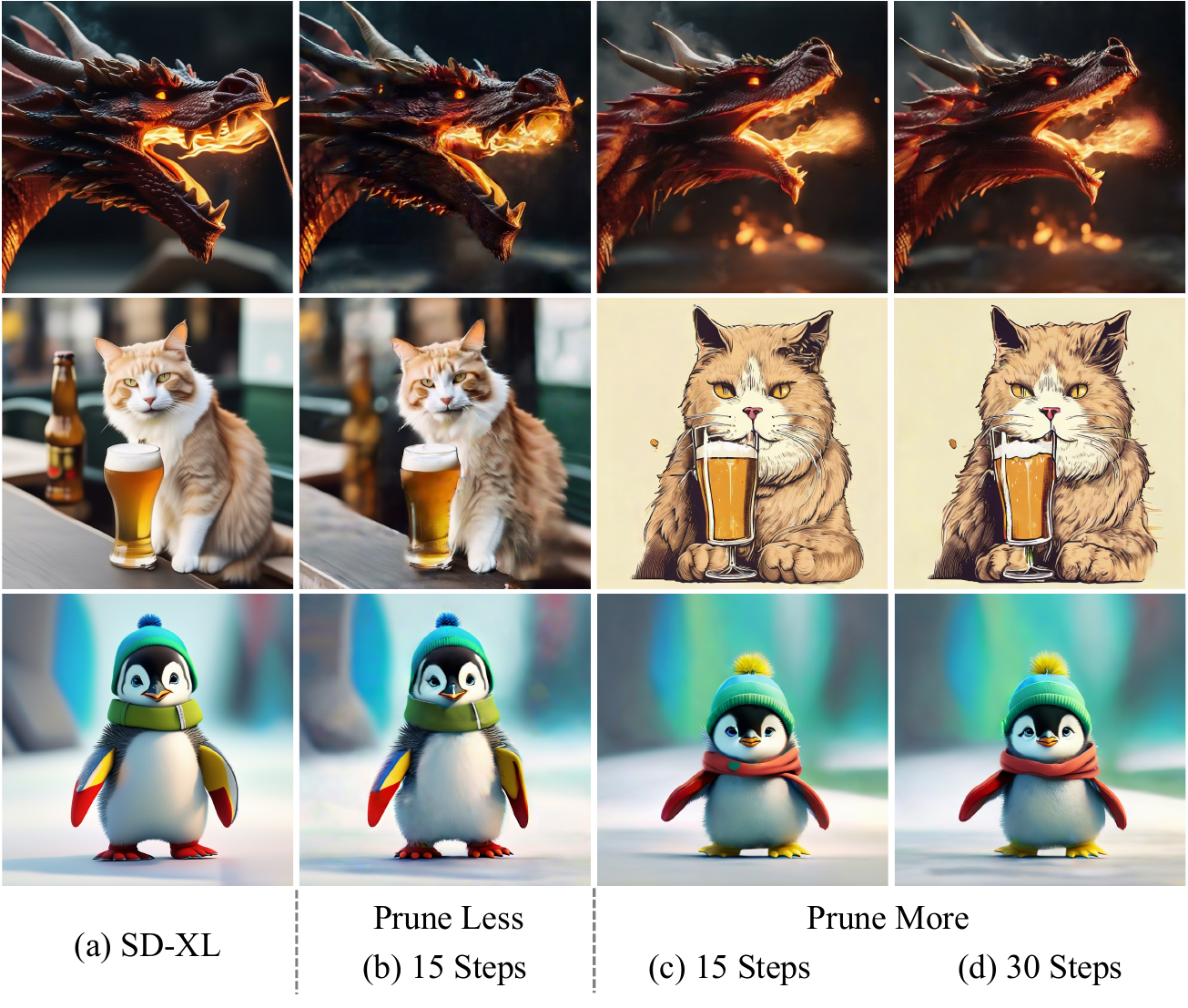}

   \caption{Comparison between different DSAP schedules. Examples indicate that pruning more tokens in early denoising steps changes the layout of generated images significantly.}
   \label{fig:comp_pmore}
\end{figure}

In AT-EDM, we design a cross-step pruning schedule that is less aggressive in early denoising steps. This is based on the intuition that (1) early denoising steps determine the layout of generated images and thus are very important; (2) the ability of early denoising steps to differentiate between unimportant tokens is weaker than that of later steps. To verify our intuition, we investigate the schedule that prunes more in early denoising steps. Note that for symmetry, \textit{``prune more in the first 15 steps"} selects corresponding attention blocks in the last 35 steps for not pruning tokens while keeping the total FLOPs cost fixed. We provide visual examples in Fig.~\ref{fig:comp_pmore} for comparison. These examples clearly support our intuition that pruning more in early denoising steps not only affects the layout of generated images but also hurts image quality.

\section{Memory Footprint of AT-EDM}
\label{app:memory}

Since we need to obtain the attention map from the first attention layer, AT-EDM cannot reduce the peak memory footprint. However, benefiting from the significantly reduced number of tokens in the following attention layers, AT-EDM reduces the average memory footprint significantly. Since PyTorch does not automatically release the redundant assigned memory when the memory requirement reduces in the later layers, we theoretically estimate the average footprint of AT-EDM, assuming the redundant occupied memory will be released in the layers with fewer tokens. We believe this is practical when the implementation is good enough. The peak and theoretical average footprint of full-size SD-XL (AT-EDM) are 19.5GB (19.5GB) and 18.8GB (12.6GB), respectively. This indicates that if we have a fine-grained pipeline schedule, \textbf{AT-EDM allows us to run 49.2\% more generation tasks} with the given VRAM restriction. 

\section{Stack with Sampling Distillation}
\label{app:stack}

Methods like consistency distillation \cite{consistency, luo2023latent} can greatly reduce the cost of DMs. Note that AT-EDM does not contradict these methods and can be deployed to speed them up further. To support this, we deploy AT-EDM in SDXL-Turbo, which is a distilled version of SD-XL. Our experimental results show that although SDXL-Turbo reduces around 95\% FLOPs cost of SD-XL, \textbf{AT-EDM can further reduce the FLOPs cost of SDXL-Turbo by 33.4\% while reducing FID by 14.5\% and only incurring 2.1\% CLIP reduction} on MSCOCO-2017. AT-EDM works as a regularizer and slightly improves the quality of images.

\section{Limitations and Trade-Offs}

\label{app:limitation}

AT-EDM demonstrates state-of-the-art results for accelerating DM inference at run-time without any retraining. 
However, as a machine learning algorithm, it inevitably has some limitations.

\noindent
(1) AT-EDM requires a pre-trained DM; since it saves computation to accelerate the model, its performance is inherently upper-bounded by the full-sized one.
While most of the time, AT-EDM matches the performance of the pre-trained model, both quantitatively and qualitatively (see experimental results in the main article), with around 40\% FLOPs reduction,   
there exist some samples where the full-sized model outperforms AT-EDM (see Fig.~\ref{fig:comp_flops}). 
Nonetheless, AT-EDM outperforms prior art by a clear margin.
In addition, AT-EDM is differentiable. We will fine-tune the pruned model to further improve quality in the future.

\noindent
(2) AT-EDM leverages the rich information stored in the attention maps, 
which could be inaccessible without incurring overhead due to the open-sourced nature of the implementation.  
For instance, SD-XL~\cite{sdxl} adopts an efficient attention library, xFormers \cite{xFormers}, 
which fuses computation to directly obtain succeeding tokens without providing intermediate attention maps. As shown in Table \ref{tab:latency}, in the case that Fused Operation (FO) is not used, using AT-EDM leads to significant latency savings. With FO under the Current Implementation (CI), AT-EDM does not result in a huge latency saving due to the cost of calculating attention maps. Reusing attention maps across steps and obtaining an approximation for them could alleviate this issue. With FO under the Desired Implementation (DI) that provides attention maps, AT-EDM's potential is fully unlocked and leads to a decent speedup.
% Therefore, we measure the theoretical inference speedup where we assume attention maps can be obtained for free in Supplementary.
% However, we expect that under another attention computation scheme that provides attention maps as intermediate tensors, 
% AT-EDM can achieve considerable speedup. 

AT-EDM is especially good at generating object-centric images, such as a portrait. It can employ a high pruning rate without hurting the main object. Generating complex scenes or tens of objects is relatively tricky for AT-EDM since it may lose some details in corner cases. In some rare corner cases where the texture details are not significant, ToMe might perform slightly better, as our algorithm may prune too many tokens in that small region. 
ToMe is indeed a strong baseline, but it is remarkable that our AT-EDM still outperforms it in most cases.

\section{Potential Negative Social Impacts}

\label{app:negimpact}

Text-to-image generative models like SD-XL have significantly advanced the field of AI and digital art creation. However, they may also potentially have negative social impact. For example, they can create highly realistic images that may be indistinguishable from real photographs. As the technology can be used to create convincing but false images, this can potentially lead to confusion and misinformation spread. In addition, the use of these models to create inappropriate or harmful content, such as realistic images of violence, hate speech, or explicit material, raises significant ethical questions. There is also the potential for perpetuating biases if the AI model is trained on biased datasets.

\end{document}